\documentclass[10pt,twocolumn,letterpaper]{article}

\usepackage{cvpr}
\usepackage{graphicx}
\usepackage{comment}
\usepackage{amsmath,amssymb} 
\usepackage{color}
\usepackage{times}
\usepackage{epsfig}

\usepackage{amsfonts}       

\newcommand{\A}{{\mathcal{A}}}

\newcommand{\be}{\begin{eqnarray}}
  \newcommand{\ee}{\end{eqnarray}}
\newcommand{\beq}{\begin{equation}\begin{aligned}}
  \newcommand{\eeq}{\end{aligned}\end{equation}}
\newcommand{\beqn}{\begin{equation*}\begin{aligned}}
  \newcommand{\eeqn}{\end{aligned}\end{equation*}}
\newcommand{\ben}{\begin{eqnarray*}}
  \newcommand{\een}{\end{eqnarray*}}

\usepackage{algorithm}
\usepackage{algorithmic}

\usepackage{multirow}
\usepackage{booktabs}       

\usepackage[pagebackref=true,breaklinks=true,letterpaper=true,colorlinks,bookmarks=false]{hyperref}

\cvprfinalcopy
\ifcvprfinal\fi

\usepackage{authblk}
\usepackage{ftnright}
\begin{document}

\title{Blind Adversarial Pruning: \\ Balance Accuracy, Efficiency and Robustness} 

\author[1]{Haidong Xie}
\author[1,2]{Lixin Qian}
\author[1]{Xueshuang Xiang\thanks{Corresponding author: xiangxueshuang@qxslab.cn}}
\author[1]{Naijin Liu}
\affil[1]{\normalsize Qian Xuesen Laboratory of Space Technology, China Academy of Space Technology}
\affil[2]{\normalsize School of Mathematics and Statistics, Wuhan University}

\maketitle

\begin{abstract}
\vspace{-0.2cm}
With the growth of interest in the attack and defense of deep neural networks, 
researchers are focusing more on the robustness of applying them to devices with limited memory. 
Thus, unlike adversarial training, which only considers the balance between accuracy and robustness, 
we come to a more meaningful and critical issue, i.e., the balance among accuracy, efficiency and robustness (AER). 
Recently, some related works focused on this issue, but with different observations, and the relations among AER remain unclear.
This paper first investigates the robustness of pruned models with different compression ratios under the gradual pruning process and concludes that the robustness of the pruned model drastically varies with different pruning processes, especially in response to attacks with large strength.
Second, we test the performance of mixing the clean data and adversarial examples (generated with a prescribed uniform budget) into the gradual pruning process, called adversarial pruning, and find the following: 
the pruned model's robustness exhibits high sensitivity to the budget, i.e., is only robust when confronting attacks with the same strength.
Furthermore, to better balance the AER, we propose an approach called blind adversarial pruning (BAP), which introduces the idea of blind adversarial training into the gradual pruning process. 
The main idea is to use a cutoff-scale strategy to adaptively estimate a nonuniform budget to modify the AEs used during pruning, thus ensuring that the strengths of AEs are dynamically located within a reasonable range at each pruning step and ultimately improving the overall AER of the pruned model. 
The experimental results obtained using BAP for pruning classification models based on several benchmarks demonstrate the competitive performance of this method: the robustness of the model pruned by BAP is more stable among varying pruning processes, and BAP exhibits better overall AER than adversarial pruning. 
\end{abstract}

\section{Introduction}

Deep learning~\cite{Hinton_Deep_learning,Goodfellow_2016} has enabled great breakthroughs in many fields, such as computer vision~\cite{Krizhevsky2012ImageNet}, speech recognition~\cite{Mikolov2012Strategies,Hinton2012Deep}, and natural language processing~\cite{Sutskever2014Sequence}. 
However, after adversarial examples (AEs) were introduced~\cite{szegedy2013intriguing,2014arXiv1412.6572G}, the weakness of deep neural networks~(DNNs) has attracted increasing attention. 
Many effective AE generation methods and defensive strategies concerning the balance between accuracy and robustness are proposed~\cite{akhtar2018threat,Zhang2018Adversarial}.  
At the same time, as the architecture of DNNs has become increasingly expressive, lightweight networks have become one of the inevitable needs. There are also many effective compression methods~\cite{Ji-Compression,CheWan17} concerning the balance between accuracy and efficiency. 
However, as DNNs may be applied on devices with limited memory and face potential attacks, we need to consider a more meaningful and critical issue, i.e., the balance among \textbf{accuracy, efficiency and robustness} (AER), which is the main goal of the proposed method. 

Of these methods to strengthen the robustness of DNNs against adversarial attacks, i.e., balancing accuracy and robustness, adversarial training (AT) is the most commonly used approach, which is a process of training a neural network on a mixture of clean data and AEs; see~\cite{szegedy2013intriguing,2014arXiv1412.6572G,madry2017towards} for white-box attacks and~\cite{kurakin2016adversarialb,tramer2017ensemble,song2018improving} for black-box attacks. 
However, for those AT approaches, it was proven numerically~\cite{madry2017towards} and theoretically~\cite{Tsipras2018} that improvement in the robustness is always accompanied by loss of accuracy. 
Furthermore, the limitation of incurring a minimal impact on accuracy requires the architecture of the network to be sufficiently expressive~\cite{akhtar2018threat,Zhang2018Adversarial}, as guaranteed by the universal approximation theorem~\cite{2014arXiv1412.6572G,hornik1989multilayer} and regularization~\cite{Sank_2017}.
Thus, these existing AT approaches may lose efficiency while improving robustness. 

Meanwhile, for balancing accuracy and efficiency, numerous advanced techniques for compacting and accelerating DNNs have been developed, such as parameter pruning~\cite{HanLearning,HASSIBI1993Second} and sharing~\cite{Chen2015Compressing}, quantization~\cite{7780890,han2015compression} and binarization~\cite{NIPS2015_5647,10.1007/978-3-319-46493-0_32}, low-rank factorization~\cite{Lebedev2015Speeding,Tai2015Convolutional}, transferred/compact convolutional filters~\cite{Cohen2016Group}, and knowledge distillation~\cite{Ba2013Do,Hinton2013Distilling}. 
Within these methods, parameter pruning is effective in reducing the network complexity and addressing the overfitting problem~\cite{CheWan17}.
Han et al.~\cite{HanLearning} proposed an unstructured parameter pruning method to reduce the total number of parameters and operations in the entire network by pruning redundant or noninformative weights in a pretrained CNN model. Although the parameter pruning method can reduce the scale of parameters without significantly decreasing the model performance (clean accuracy), it is unable to ensure the robustness of the model, as discussed in Reference~\cite{5989836} and Section~\ref{sec:RvsR}. 

\begin{figure*}[t]
	\centering
	\includegraphics[width=0.3\linewidth]{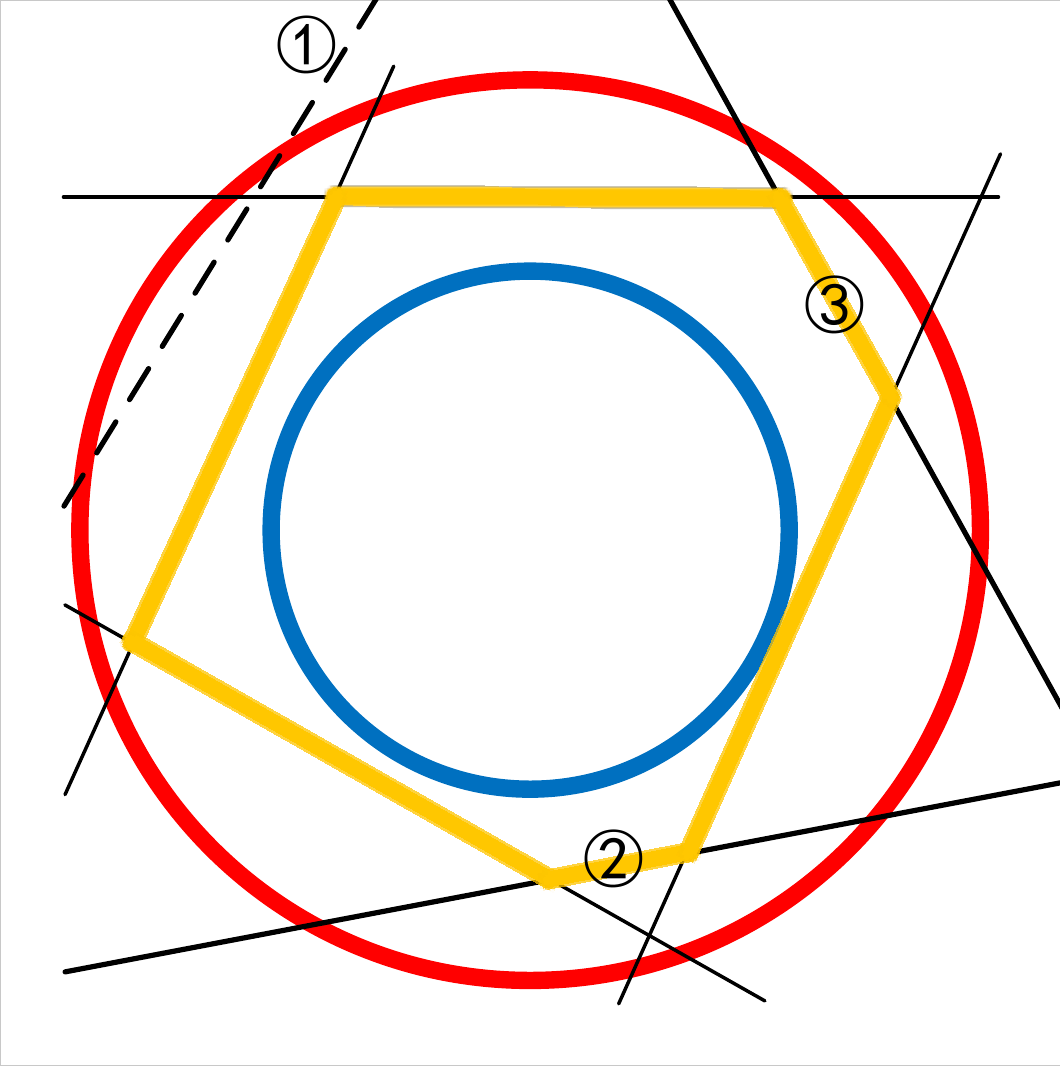}
	\includegraphics[width=0.3\linewidth]{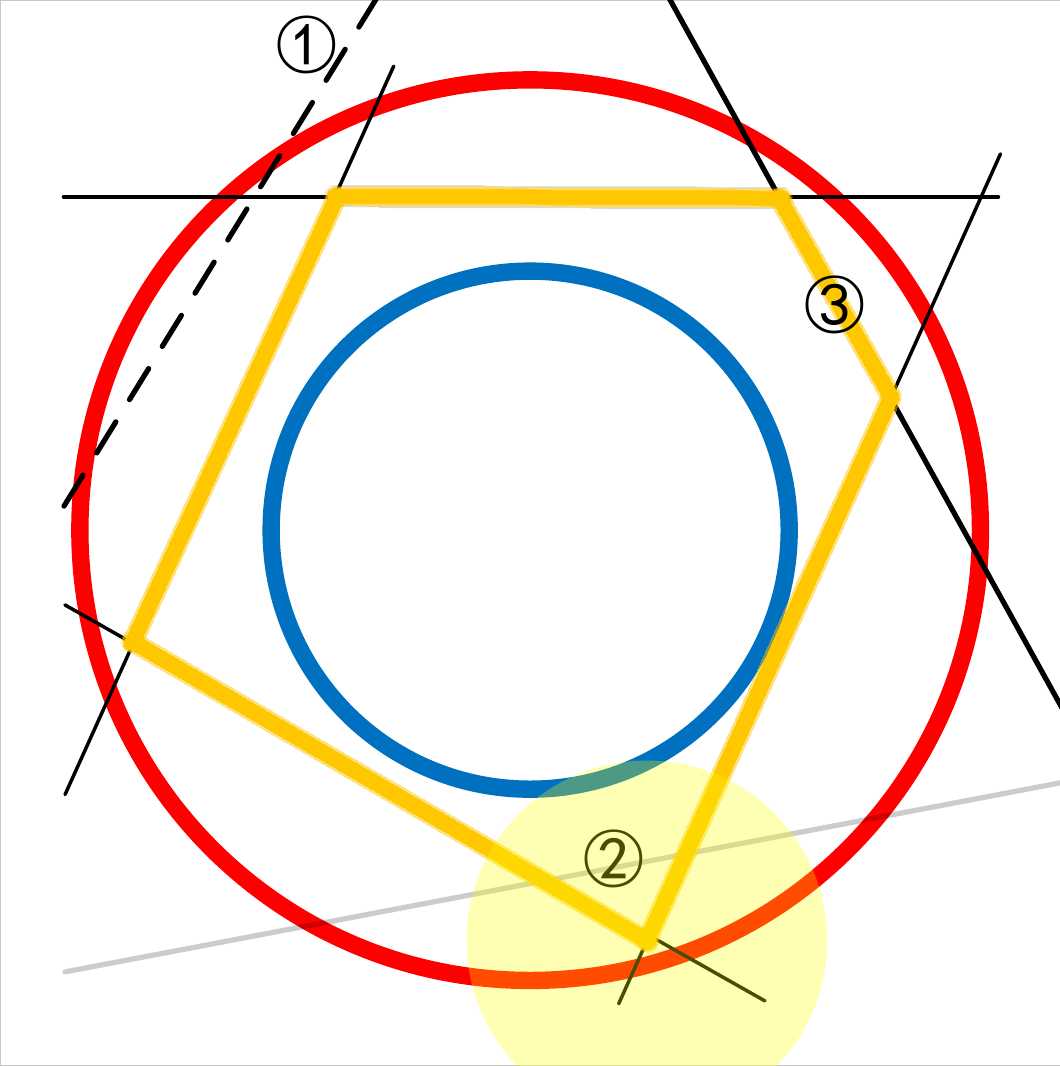}
	\includegraphics[width=0.3\linewidth]{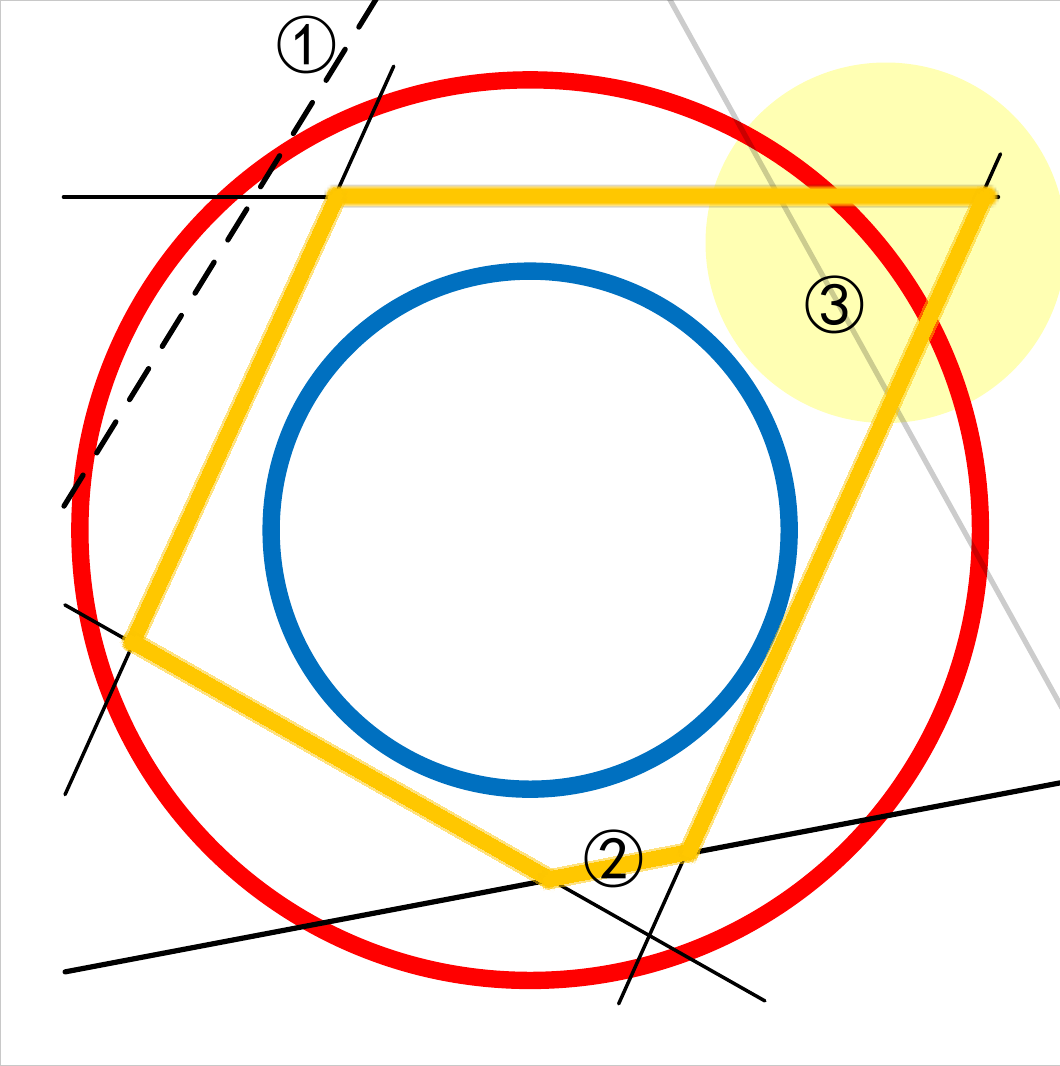}
	\caption{Schematic diagram of the two circles classification problem with model pruning. Red and blue circles correspond to the data distribution to be classified. The orange polygon corresponds to the decision boundary of the training model. Edges 1, 2 and 3 represent different kinds of edges that may be pruned. 
		(Left) Pruning edge 1 will have no impact on the accuracy and robustness of the model. 
		(Middle) Pruning edge 2 will have no impact on the accuracy but will reduce the robustness of the model. 
		(Right) Pruning edge 3 will reduce both accuracy and robustness of the model. }
	\label{fig:tcc}
\end{figure*}

Therefore, the exploration of the relationship between AER of DNNs requires more attention. There are few related research efforts focused on this stage, and the conclusions are divergent. Some of the studies indicate that adversarial robustness requires a significantly larger capacity of the network than that for the normal training while maintaining accuracy both numerically~\cite{madry2017towards} and theoretically~\cite{Tsipras2018,DBLP:journals/corr/abs-1901-00532}. The others suggest active pruning as a defense method~\cite{DBLP:journals/corr/abs-1904-08444,DBLP:journals/corr/abs-1803-01442}. References~\cite{YiwenSparse,Kai18Training} theoretically analyze the relationship between efficiency and robustness; their results show that a proper compression ratio can increase the robustness of the model. Guo et al.~\cite{YiwenSparse} state that only the pruned model with less than $2.5\%$ nonzero weights can achieve the best robustness under FGSM attacks with $\ell_\infty$ norm $\varepsilon=0.1$.
Furthermore, different pruning ratios exert a drastic nonmonotonic impact on robustness, and appropriate sparsity can lead to both good accuracy and robustness, whereas over-sparsification may cause model fragility; the inflection point is related to the attack budgets. 
We now give a simple explanation of how these conclusions diverge. 
It is known that the main strategy of model compression, such as model pruning, is to remove or compress redundant weights of the loss-function involving only the clean data. 
Thus, intuitively, normal training with model pruning (denoted as \textbf{normal pruning (NP)}) discards weights based on the clean data during pruning, 
which can ensure good performance on the clean data but cannot guarantee the robustness with respect to adversarial attacks. 
We can qualitatively analyze this by a simple two circles classification problem as in Figure~\ref{fig:tcc}. 
If we use NP, the pruning only takes into account the loss of clean data. 
Discarding the weights corresponding to edges 1 and 2 will not lead to a decrease in clean accuracy: obviously, only pruning edge 1 will exert no impact on the robustness of the model, while cutting edge 2 will lower the robustness instead. 
Since the importance of edges 1 and 2 to the robustness is not reflected in the loss, we may expect that the pruning process will prune one of them randomly. 
This means that if we repeat pruning several times independently, we will find that the model's robustness fluctuates greatly. 
More numerical experiments and discussions about the robustness of pruned models with different compression ratios under the gradual pruning process (training a random initial model to one with high accuracy; then, gradually pruning it to one with a target compression ratio) will be carried out later, and it can be concluded that model pruning cannot ensure good robustness of the pruned model. Thus, \textbf{NP exhibits great fluctuation or instability with different pruning processes}, especially under attacks with large strength.

There also exist few related works focused on how to better balance AER. 
References~\cite{DBLP:journals/corr/abs-1903-12561} combine adversarial training and weight pruning, finding that the pruning of a large model can achieve both higher clean accuracy and adversarial robustness than a small model. ATMC proposed by Gui et al.~\cite{DBLP:journals/corr/abs-1902-03538} obtains a remarkably more favorable trade-off among model size, accuracy and robustness. 
{Sehwag et al.~\cite{Sehwag2020PruningAdversarially} improve the robustness of the pruning model by introducing scaled-initialization of the importance scores in each layer.}
Their results could be explained according to the benefits of overparameterization~\cite{DBLP:journals/corr/abs-1811-08888,DBLP:journals/corr/abs-1811-04918,DBLP:journals/corr/abs-1811-03962}, and they focus on improving robustness for a given efficiency or improving efficiency without loss of robustness, respectively. 
We refer to this type of method combining adversarial training and model pruning as an \textbf{adversarial pruning (AP)} method. 
We should notice that a primary characteristic which the existing AP methods have in common is to use norm-constrained AEs with a prescribed uniform budget in the gradual pruning process, and we evaluate the robustness of models by AEs with the same budget. 
Because the model discarded weights based on adversarial loss-function during pruning, it can ensure good performance on the trained data (the clean data and AEs of the prescribed budget), but it cannot guarantee the robustness of attacks with other budgets. 
From the numerical experiments and discussions about AP, we can conclude that \textbf{the pruned model's robustness shows high sensitivity to the budget by AP}; i.e., the pruned model obtained with the prescribed uniform budget is only robust when confronting attacks with the same strength. 

To better analyze the relation between AER, we first propose an evaluation criterion, also called \textbf{AER}, which can comprehensively describe the comprehensive performance of the algorithm with respect to accuracy, efficiency and robustness. 
For the later discussion on AER of NP and AP, we can clearly observe the abovementioned drawbacks of NP and AP.
Furthermore, to ameliorate these drawbacks, i.e., better balance the AER, we propose an approach called \textbf{blind adversarial pruning (BAP)} which introduces the idea of blind adversarial training~\cite{BAT} into the gradual pruning process~\cite{han2015compression}. 
Like BAT, the main idea of BAP is to use a cutoff-scale strategy to adaptively estimate a nonuniform budget to modify the AEs used in the gradual pruning, thereby ensuring that the strengths of the AEs are dynamically located within a reasonable range at each pruning step and ultimately improving the overall AER of the pruned model. 
The experimental results obtained using BAP for pruning classification models based on several benchmarks (LeNet-$5$ on MNIST, and FitNet-$4$ on CIFAR10 and CIFAR100) demonstrate the competitive performance of this method: the robustness of the pruned model by BAP is more stable with the varying pruning process; BAP exhibits better overall AER than NP and AP. 

\section{Analysis of AER} \label{sec:RvsR}

As discussed above, the relationship among accuracy, efficiency and robustness (AER) seems to be highly non-straightforward and contextually varying; this section explores this relationship by focusing on efficiency with respect to model pruning and robustness under adversarial attacks. Thus, we use the {compression ratio}~\cite{CheWan17} and {attack budget of AEs (also called norm of AEs perturbation)}~\cite{2014arXiv1412.6572G} to measure the model pruning and adversarial attack. 

$M_{\alpha}$ denotes the compression model of DNN model $M$ with \textbf{compression ratio} $\alpha \in (0,1]$. 
This means that $\alpha = |M_{\alpha}| / |M|$, the ratio of the number of unpruned weights to the total weights. 
For given AEs under $\ell_p$-norm, the \textbf{attack budget} $\varepsilon$ can be defined as
\begin{equation}
\varepsilon = || \delta(\mathbf{x}) ||_p = || \mathbf{x} - \mathbf{x}_\text{Adv} ||_p,
\label{eq:norm}
\end{equation}
where $\delta(\mathbf{x})$ is the adversarial perturbation, $\mathbf{x}$ and $\mathbf{x}_\text{Adv}$ correspond to the clean data and AEs, and we set $p=\infty$ for FGSM~\cite{2014arXiv1412.6572G} and $p=2$ for DeepFool~\cite{moosavi2016deepfool}.

Existing methods only change the compression ratio or attack budget alone; for example, reference~\cite{DBLP:journals/corr/abs-1903-12561} compares different algorithms by considering a given compression ratio and attack budget, while reference~\cite{DBLP:journals/corr/abs-1902-03538} compares different algorithms by observing different compression ratios in a fixed attack budget. These methods cannot give comprehensive information about AER. 
To comprehensively characterize AER, this paper considers traversing compression ratio and attack budget to comprehensively evaluate the training algorithms. We refer to our evaluation criterion as \textbf{AER (accuracy, efficiency and robustness)}, which is defined by
\begin{equation}
\text{AER}(\alpha, \varepsilon) = \A(\mathbf{x} + \delta(\mathbf{x}, \varepsilon) , M_{\alpha})
\label{eq:aer}
\end{equation}
where $\varepsilon$ donates the attack budget as Equation~\eqref{eq:norm}, and $\alpha$ is the compression ratio. $\A(\cdot)$ shows the accuracy of model $M_{\alpha}$ with data $\mathbf{x} + \delta(\mathbf{x}, \varepsilon)$, and $\delta(\mathbf{x}, \varepsilon)$ means cutoff by the norm of adversarial perturbation $\delta(\mathbf{x})$ with budget $\varepsilon$,
which is defined as $\delta(\mathbf{x},\varepsilon)$: if $||\delta(\mathbf{x})|| > \varepsilon$, $\delta(\mathbf{x}) \leftarrow \delta(\mathbf{x}) / ||\delta(\mathbf{x})|| \cdot \varepsilon$; otherwise, there is no change in $\delta(\mathbf{x})$.
It can be observed by definition that AER can describe the comprehensive performance of the algorithms with respect to accuracy, efficiency and robustness.

Furthermore, to simplify the expression and to exhibit improvements more intuitively in Section~\ref{sec:Result}, \textbf{average AER} is defined to show the overall performance within an interval of compression ratio $\alpha$ and attack strength $\varepsilon$ according to the expectation of the collection of several independent pruning processes $\xi$,
\begin{equation}
\text{avg-AER}(\Gamma, \Delta) = \mathbb{E}_{\xi} \frac{1}{1-\Gamma} \frac{1}{\Delta} \int_{\Gamma}^{1} \int_{0}^{\Delta}
\text{AER}(\alpha, \varepsilon) d \alpha d \varepsilon.
\label{eq:a-aer}
\end{equation}
In actual calculations, for each pruning process, we first numerically compute the two-dimensional integration about $\alpha$ and $\varepsilon$ and then average these values for several pruning processes.
A larger $\text{avg-AER}(\Gamma, \Delta)$ means that the model exhibits greater overall performance.

\section{Blind Adversarial Pruning} \label{sec:Method}

The comprehensive performance of the models obtained by different training strategies is completely different, and introducing the idea of adversarial training can significantly improve the robustness of the model under the same compression ratio. References~\cite{DBLP:journals/corr/abs-1903-12561} and~\cite{DBLP:journals/corr/abs-1902-03538} combine model pruning and adversarial training into a unified optimization framework, but they only consider restricted AT and the trained model is more capable of resisting attacks of the given budget. Thus, they only improve the performance of $\text{AER}(\alpha, \varepsilon)$ with a fixed attack budget $\varepsilon$, and this cannot effectively represent the comprehensive performance.

Here, we combine the ideas of blind adversarial training (BAT) and model pruning, proposing \textbf{blind adversarial pruning (BAP)}, as shown formally in the following formula,
\begin{eqnarray}
\min_\theta \mathbb{E}_{(\mathbf{x},\mathbf{y})} \max_{\delta(\mathbf{x})} \{\mathcal{L}(\theta,  \underbrace{\mathbf{x}+\rho \delta(\mathbf{x})}_{\rm Scale},\mathbf{y}) 
- \lambda_1||\delta(\mathbf{x})||_2 \nonumber\\
- \underbrace{\lambda_2(||\delta(\mathbf{x})||_2 - \varepsilon)_{+}}_{\rm Cutoff}\} 
+ \underbrace{\lambda_3(||\theta||_0 - \alpha|\theta|)_+}_{\rm Pruning}. 
\label{eq:opt_bat}
\end{eqnarray}
BAP consists of three parts: `cutoff' and `scale' form the BAT part, and `pruning' forms the model pruning part.
Within the BAT part, the two parameters $\varepsilon$ and $\rho$ are used to control the cutoff and the scale process, respectively. The parameter $\alpha$ corresponds to the targeted compression ratio, and $|\theta|$ indicates the number of elements.
Values of $\lambda_i,i=1,2,3$ represent the Lagrangian multiplier coefficient, and the formula in the calculation will be solved alternately in blocks, and so the coefficients are only formal parameters.

The procedure of BAP is given in algorithm~\ref{alg:bat-p}, with BAT and pruning parts. 
BAT attempts to produce a model that can fully resist attacks of different budgets~\cite{BAT} by using the unrestricted AE method, DeepFool, which aims for the decision boundary of a model. BAT proposes a {cutoff-scale}~(CoS) strategy based on the DeepFool-AT in order to ensure that the AEs are dynamically located within a reasonable range such that the AT model can be robust when encountering attacks of varying strengths. 
The BAT part starting with the AEs generated by DeepFool corresponds to only maximizing the first two terms in the objective of Equation~\eqref{eq:opt_bat} (with $\rho=1$), and we simply calculate the third term (equivalent to minimizing $(||\delta(\mathbf{x})|| - \varepsilon)_{+}$) by cutting off the perturbations of AEs with a norm larger than $\varepsilon$, as defined by $\delta(\mathbf{x},\varepsilon)$: if $||\delta(\mathbf{x})|| > \varepsilon$, $\delta(\mathbf{x}) \leftarrow \delta(\mathbf{x}) / ||\delta(\mathbf{x})|| \cdot \varepsilon$; otherwise, no change of $\delta(\mathbf{x})$.
Then, we scale the new perturbations with weight $\rho$ and add these CoS AEs into the training process, i.e., $\{\mathbf{x}+\rho \delta(\mathbf{x})\}$. 
We set $\varepsilon= \mathbb{E}||\delta(\mathbf{x})||$ and $\rho<1$ (a predefined parameter), corresponding to adaptively estimating a nonuniform budget. 
$\delta(\mathbf{x},\mathbb{E}||\delta(\mathbf{x})||)$ implies that the budget $\varepsilon\leftarrow \mathbb{E}||\delta(\mathbf{x})||$ is computed prior to cutting off the perturbations $\delta(\mathbf{x})$. 
For more details, see the BAT paper~\cite{BAT}.

Next, pruning is conducted by pruning the model parameters according to their importance, and the redundant parameters of the expected proportion are gradually converted to 0~\cite{han2015compression}. 
First, we obtain the current compression ratio $\alpha_\text{step} = \alpha + (1-\alpha) * (1 - {j_\text{step}}/{(N_\text{epoch}N_{\text{step per epoch}})}) ^ 3$, where $\alpha$ is the targeted compression ratio, $j_\text{step}$ indicates the current step number, $N_\text{epoch}$ denotes the number of epochs that reach targeted $\alpha$ and $N_{\text{step per epoch}}$ means the number of steps per epoch. Then, we sort model parameters $\theta_i$ by importance and set redundant parameters to $\theta_i = 0$ until reaching the current compression ratio $\alpha_\text{step}$ if $i<\alpha_\text{step} |\theta|$, where $|\theta|$ is the total number of parameters.
Alternately, these two parts update $\theta$ gradually if $\theta \neq 0$ until the model reaches the expected compression ratio and the accuracy values of Cos AEs converge to the highest value. 

\begin{algorithm}[t]
  \caption{Blind Adversarial Pruning (BAP)}
  \label{alg:bat-p}
  \begin{algorithmic}
    \REQUIRE {Dataset $\{\mathbf{x},\mathbf{y}\}$ and hyper-parameters~(scale factor $\rho$, comparession ratio $\alpha$, learning rate $\beta$).}
    \ENSURE {Model with weights $\theta$.}
    \STATE {Initialize model weights $\theta$.}
    \REPEAT
    \STATE {---\textbf{BAT part}---}
    \STATE {$\mathcal{L}_{\rm C} = \mathcal{L}(\theta,\mathbf{x},\mathbf{y})$,} \COMMENT{Loss on clean data}
    \STATE {$\delta(\mathbf{x}) = \mathbf{x}_{\rm adv} - \mathbf{x}$,} \COMMENT{$\mathbf{x}_{\rm adv}$: DeepFool AEs}
    \STATE {$\varepsilon = \mathbb{E}_{\mathbf{x}}\,||\delta(\mathbf{x})||_2$,} \COMMENT{Adaptive Cutoff budget}
    \STATE {$\delta_{\rm Co}(\mathbf{x}) = \text{cut}\{\delta(\mathbf{x}), \varepsilon\}$,} \COMMENT{Cutoff AEs}
    \STATE {$\delta_{\rm CoS}(\mathbf{x}) = \rho \delta_{\rm Co}(\mathbf{x})$,} \COMMENT{Scale AEs}
    \STATE {$\mathbf{x}_{\rm CoS} = \mathbf{x} + \delta_{\rm CoS}(\mathbf{x})$,} \COMMENT{Get CoS AEs}
    \STATE {$\mathcal{L}_{\rm AE} = \mathcal{L}(\theta,\mathbf{x}_{\rm CoS},\mathbf{y}) $,} \COMMENT{Loss on CoS AEs}
    \STATE if $\theta_i \neq 0$, {$\theta = \theta - \beta (\nabla_{\theta} \mathcal{L}_{\rm C} + \nabla_{\theta} \mathcal{L}_{\rm AE})$,} 
        \COMMENT{Update with total loss}
    \STATE {---\textbf{Pruning part}---} 
    \STATE {$\alpha_\text{step} = \alpha + (1-\alpha) * (1 - {j_\text{step}}/{(N_\text{epoch}N_{\text{step per epoch}})}) ^ 3$} \COMMENT{Current $\alpha$}
    \STATE {Sort $\theta_i$ in ascending order of importance} \COMMENT{Order of importance}
    \STATE {$\theta_i = 0$ if $i<\alpha_\text{step} |\theta|$} \COMMENT{Prune}
    \UNTIL{Reach $\alpha$ and accuracy of Cos AEs converge.} \COMMENT{comprehensive performance}
  \end{algorithmic}
\end{algorithm}

The proposed BAP method can obtain a model with comprehensive robustness for a given compression ratio.
Compared with other training methods, the model obtained by BAP exhibits better comprehensive performance including accuracy, efficiency and robustness. See Section~\ref{sec:Result} for more information.

\section{Experimental Results} \label{sec:Result}

In this section, we evaluate the BAP approach on various benchmark datasets.
The code for these experiments is based on the open source library cleverhans~\cite{papernot2018cleverhans} for adversarial attack and tensorflow/model-optimization~\cite{model-optimization} for model pruning.
We consider using BAP to train LeNet-$5$~\cite{MnistLeNet} for MNIST~\cite{MnistLeNet} and train FitNet-$4$~\cite{Romero2015} for CIFAR-10 and CIFAR-100~\cite{Krizhevsky2009}. We compare the BAP approach with several state-of-the-art training approaches, such as normal training with model pruning (NP), FGSM-AT (AT with FGSM AEs) with model pruning (AP (FGSM)), and DF-AT (AT with DeepFool AEs) with model pruning (AP (DeepFool)). 
We mark the type of AE used in the brackets behind the AP to distinguish between different algorithms, where AP (FGSM) is also accompanied by a budget mark, such as AP (FGSM Budget 0.1) as in Figure~\ref{fig:AT-FGSM}.

\paragraph{\bf Experimental setup.} For all of the experiments, we normalize the pixels to $[0,1]$ by dividing by 255, use label smoothing regularization~\cite{7780677} to avoid overfitting, and perform data augmentation (with a width/height shift range of $0.1$ and random flips) for the CIFAR-10 and CIFAR-100 datasets to improve the clean data accuracy. 
For AE generations for all of the datasets, we use norm-constrained FGSM~\cite{2014arXiv1412.6572G} AEs with $\ell_\infty$ norm, and unconstrained DeepFool~\cite{moosavi2016deepfool} AEs with $\ell_2$ norm (number of steps: $5$ during training and $100$ during evaluation).  

For the training process, we use Adam optimization~\cite{Kingma2014Adam} with batch size $128$ and set the epochs/learning rate of MNIST as 70/1e-3, with those of CIFAR-10 and CIFAR-100 set as 200/1e-4 (NP) and 300/1e-4 (AP \& BAP), respectively.
We set the scale parameter $\rho = 0.9$ in the BAT part of BAP for the final results. 
Within the pruning part, we prune the model beginning at epoch $0$ with pruning frequency $100$ batches/steps, ending at epoch $30$ for MNIST and epoch $150$ for CIFAR-10 and CIFAR-100; see algorithm~\ref{alg:bat-p} for the dynamic variation formula of compression ratio.

\paragraph{\bf Compute Overhead.} BAP combines the characteristics of BAT and model pruning, so the calculation cost of each step is the superposition of these two parts, with no additional calculation increase. 
From the perspective of the number of epochs, the requirements of BAP and AP methods are essentially the same, and the numerical experimentation shows that slightly more epochs are required than for model pruning with NT.
Therefore, the BAP method is very efficient in terms of calculation cost.

Unlike existing work, this article proposes to show the average and standard deviation of the results through repeated experiments. This approach has resulted in large amounts of calculation, especially for CIFAR-10 and CIFAR-100 datasets. 
Therefore, in consideration of the limited computing resources, we have repeated the calculation results in a targeted manner and indicated them in detail in the caption of each image. The existing results can fully explain the improvement effect of our proposed BAP algorithm.

\subsection{AER of NP and AP under FGSM attack}

\begin{figure*}[t]
  \centering
  \includegraphics[width=0.3\linewidth]{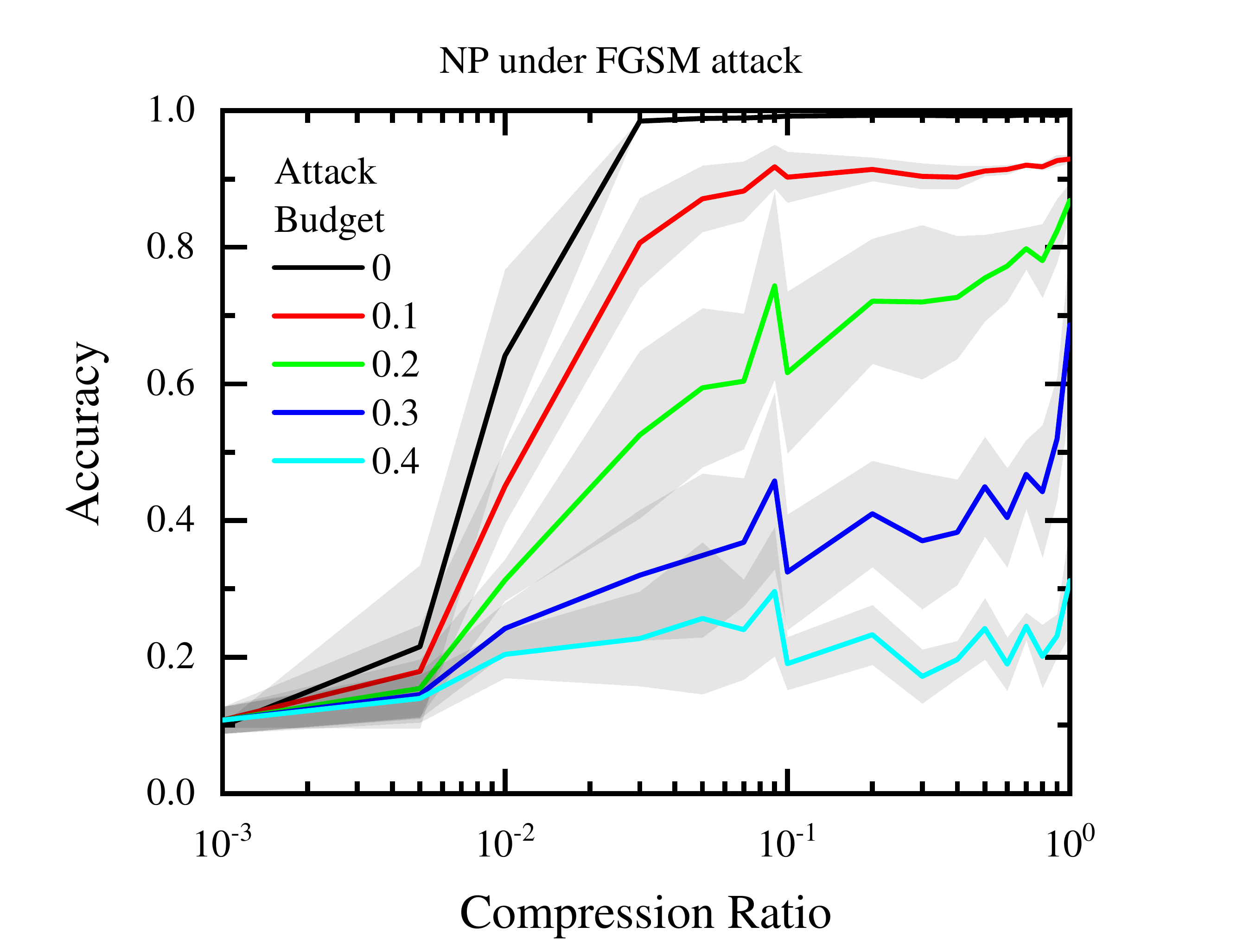}
  \includegraphics[width=0.3\linewidth]{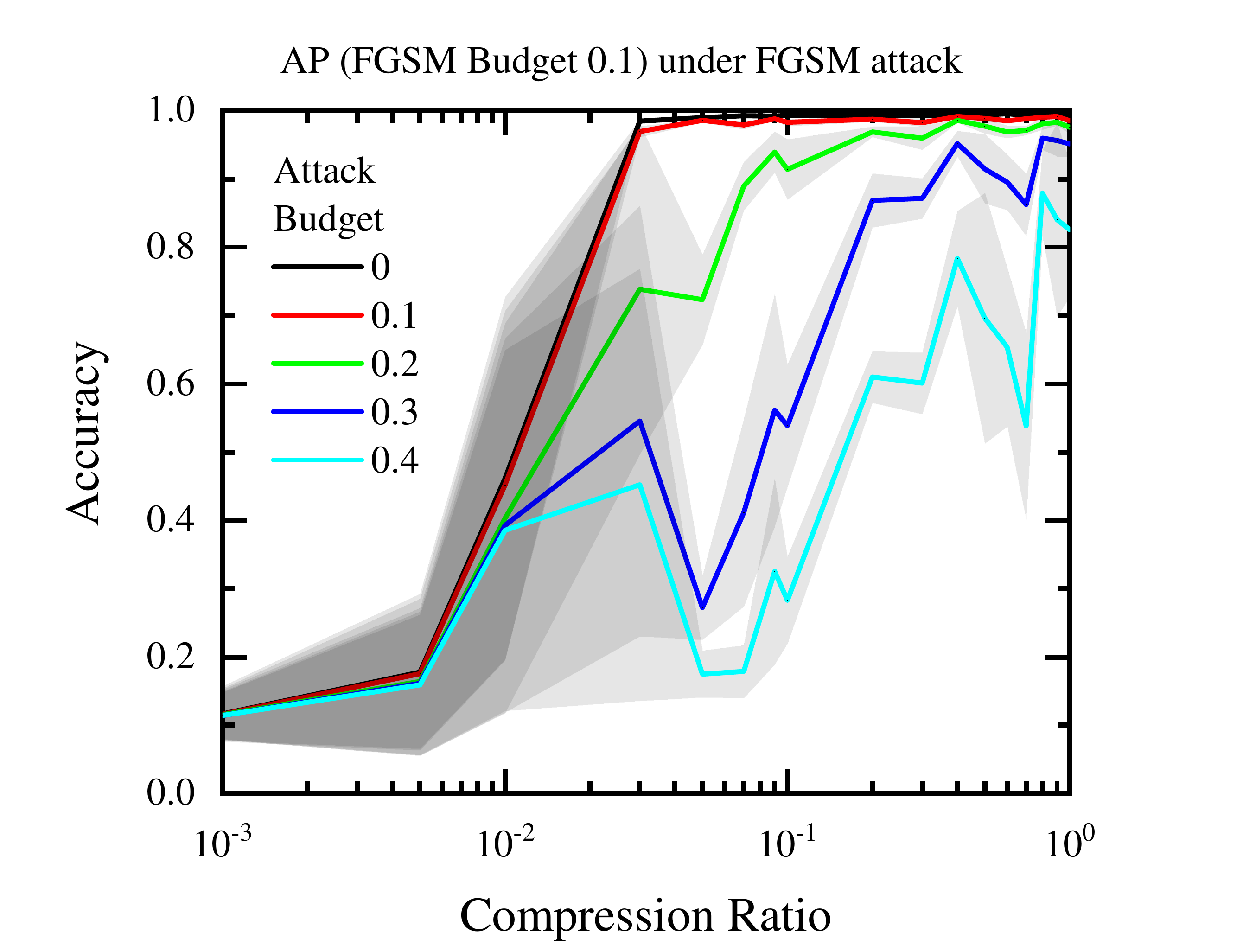}
  \includegraphics[width=0.3\linewidth]{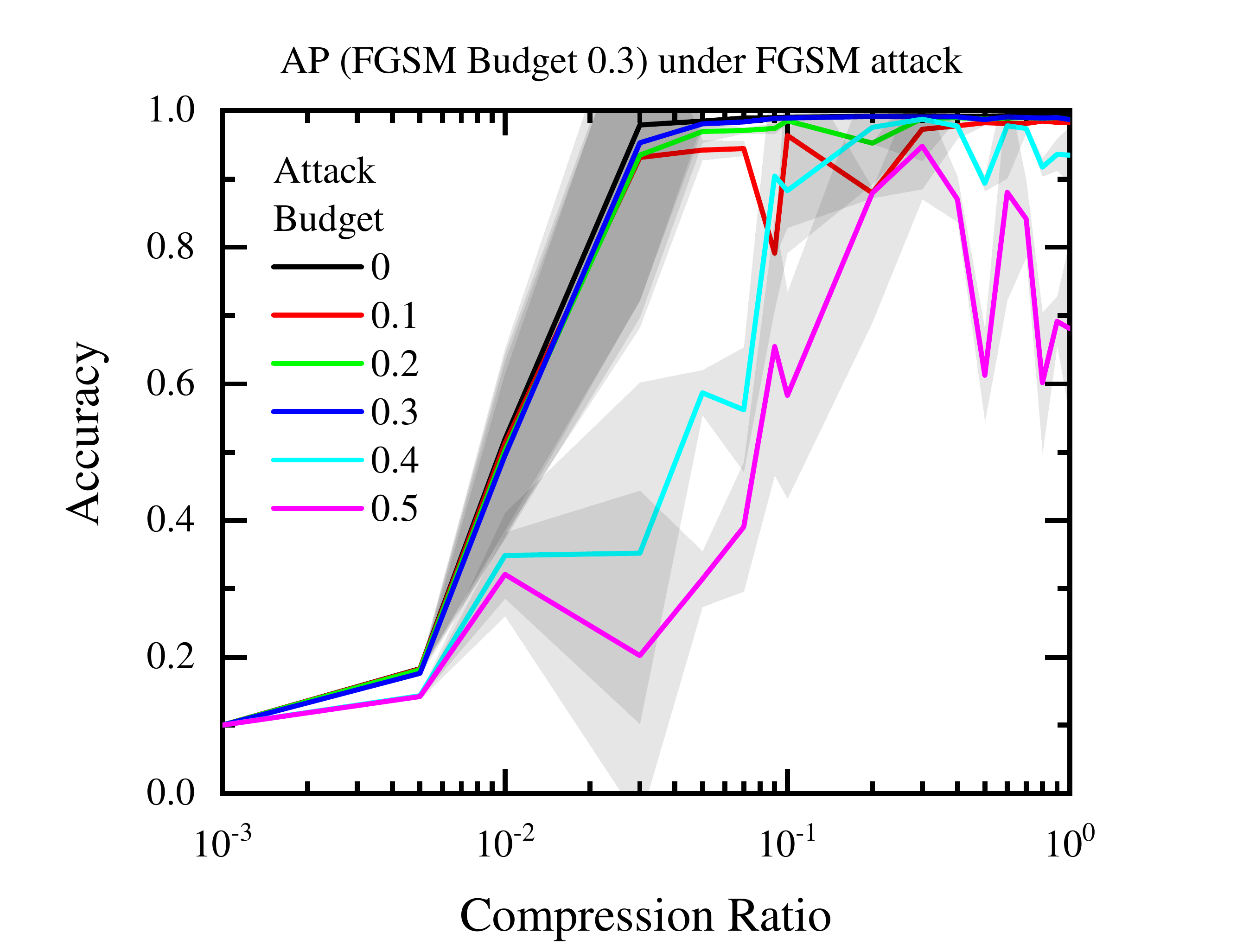}
  \includegraphics[width=0.3\linewidth]{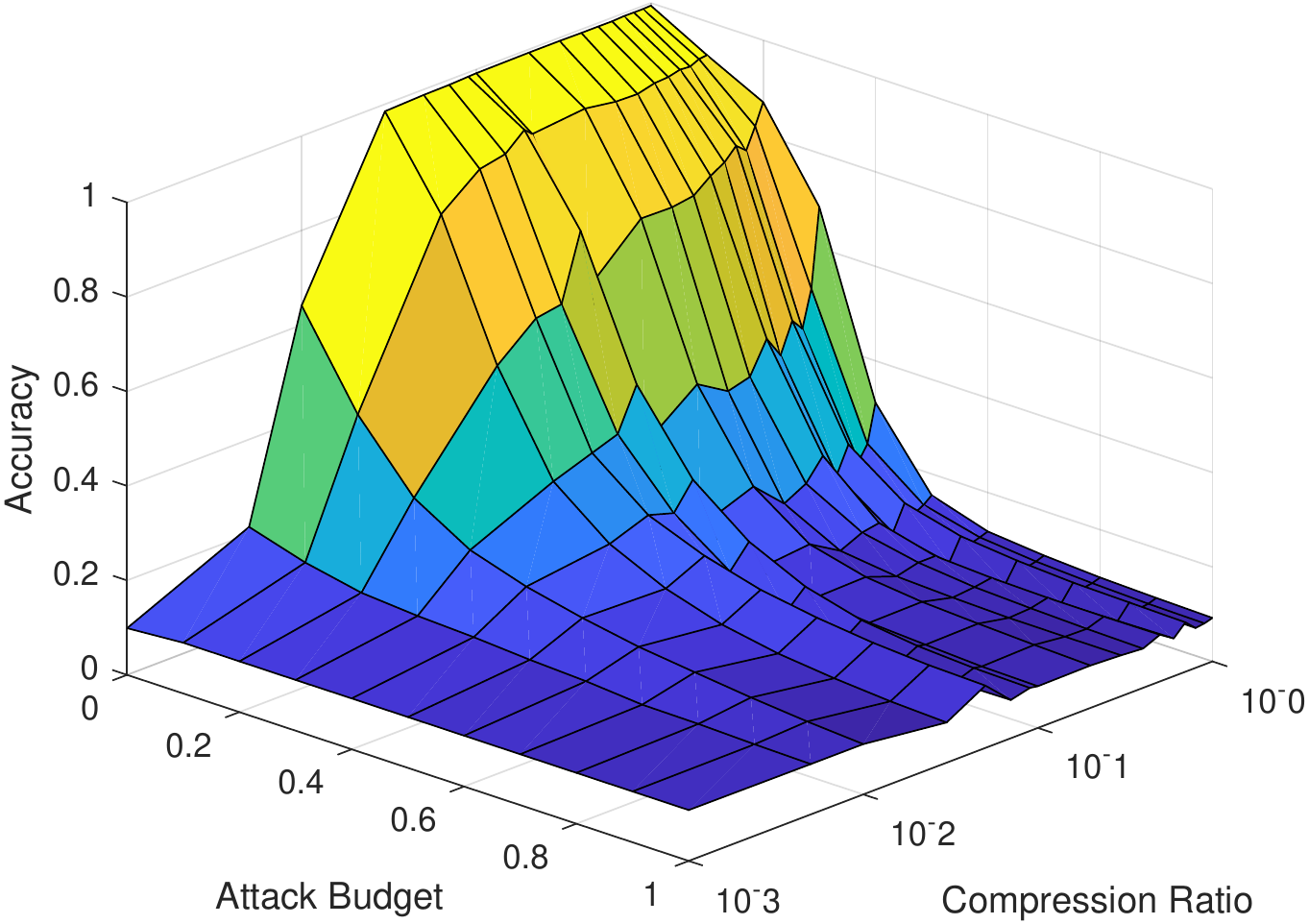}
  \includegraphics[width=0.3\linewidth]{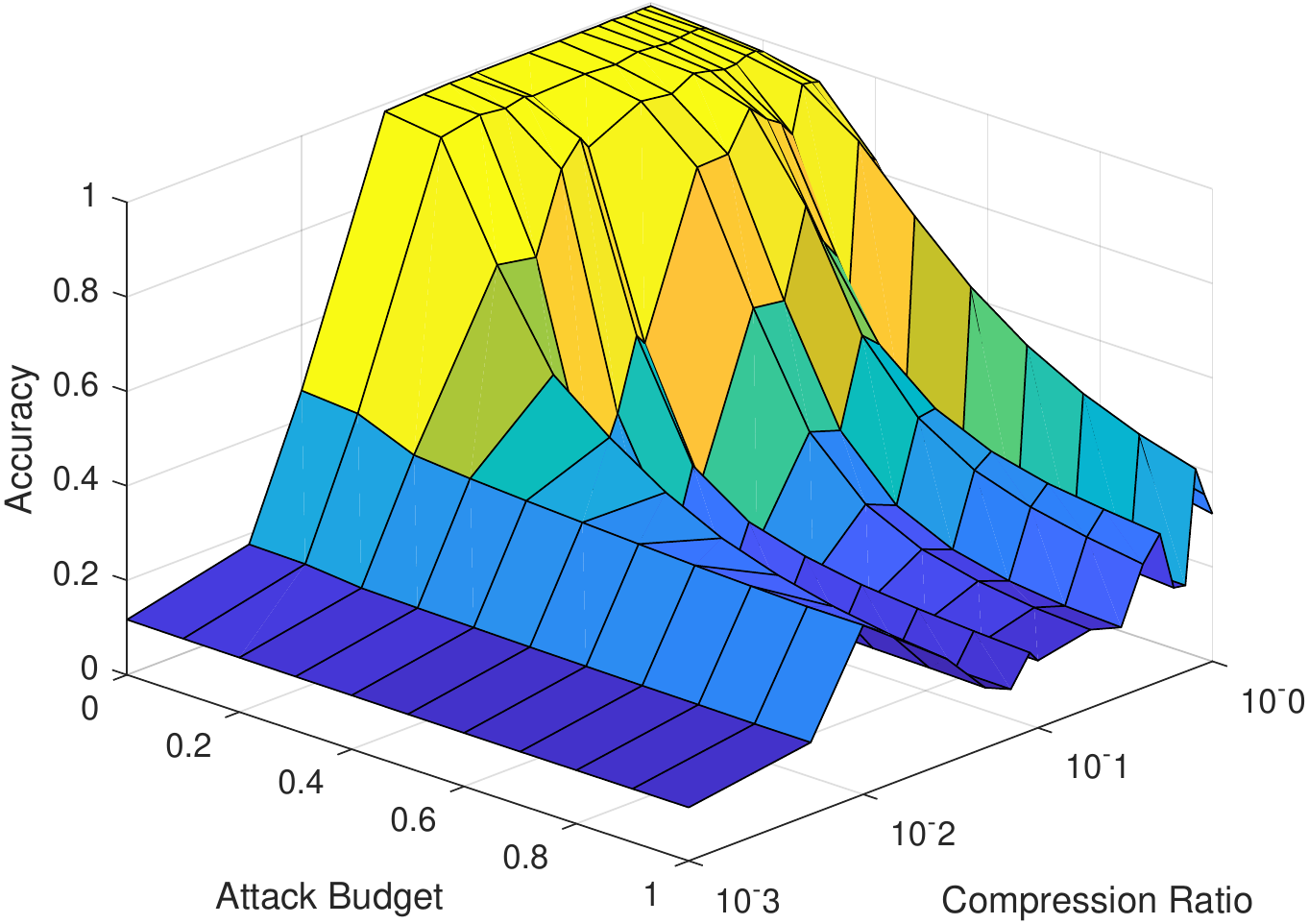}
  \includegraphics[width=0.3\linewidth]{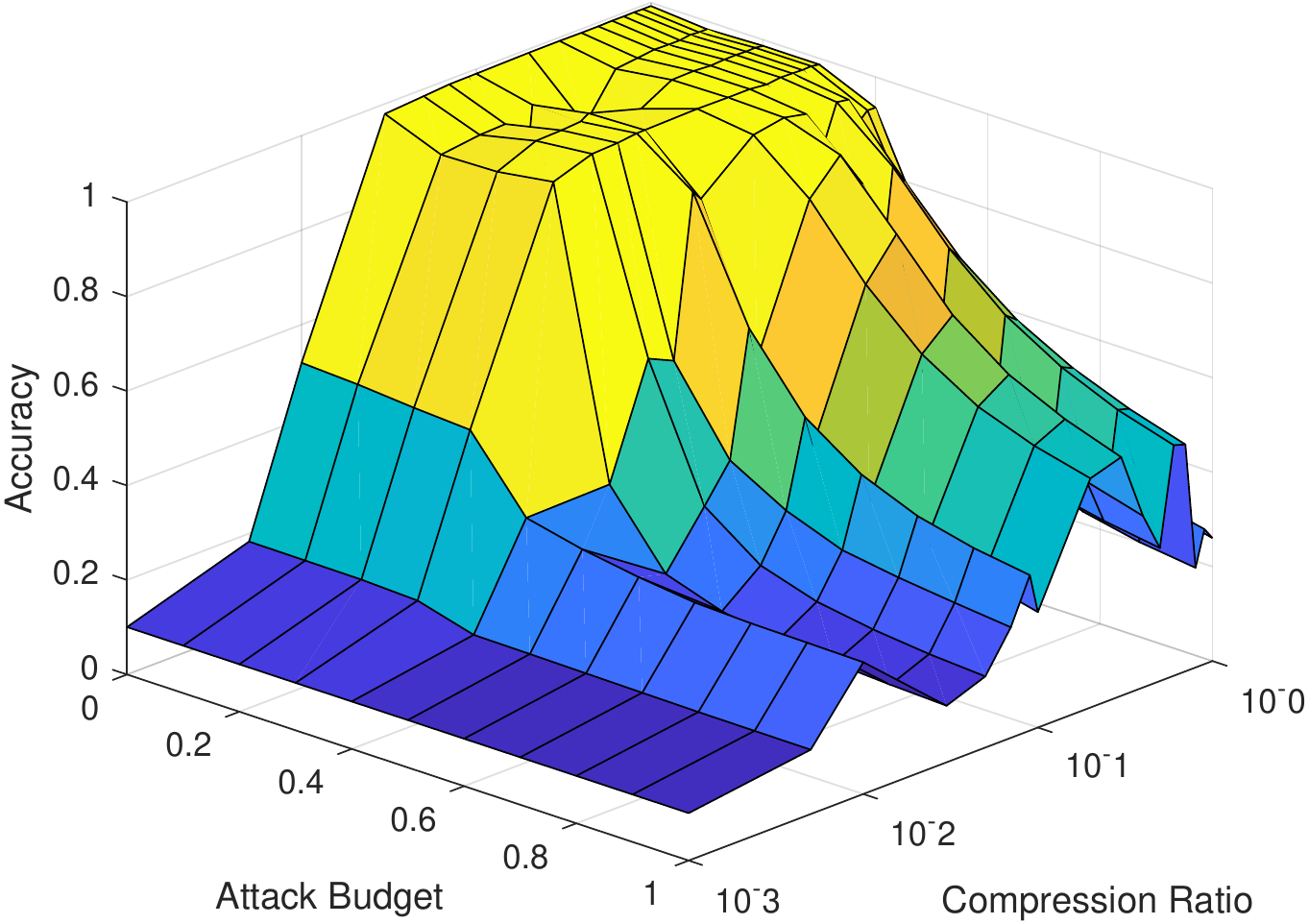}
  \caption{Comparison between the NP model and AP model (FGSM budgets $\varepsilon=0.1 \& 0.3$) of different compression ratios and different attack budgets for \textbf{MNIST} datasets. 
  Each column corresponds to a different training method, while in each column: 
  (Upper) The solid colored lines and the corresponding gray area represent the average and standard deviation of the accuracy with different attack budgets obtained by independently repeating the pruning process 5 times. 
  The wider the gray area is, the more unstable the model’s robustness.
  (Lower) Three-dimensional representation of AER.} 
  \label{fig:AT-FGSM}
\end{figure*}

\begin{figure*}[t]
  \centering
  \includegraphics[width=0.3\linewidth]{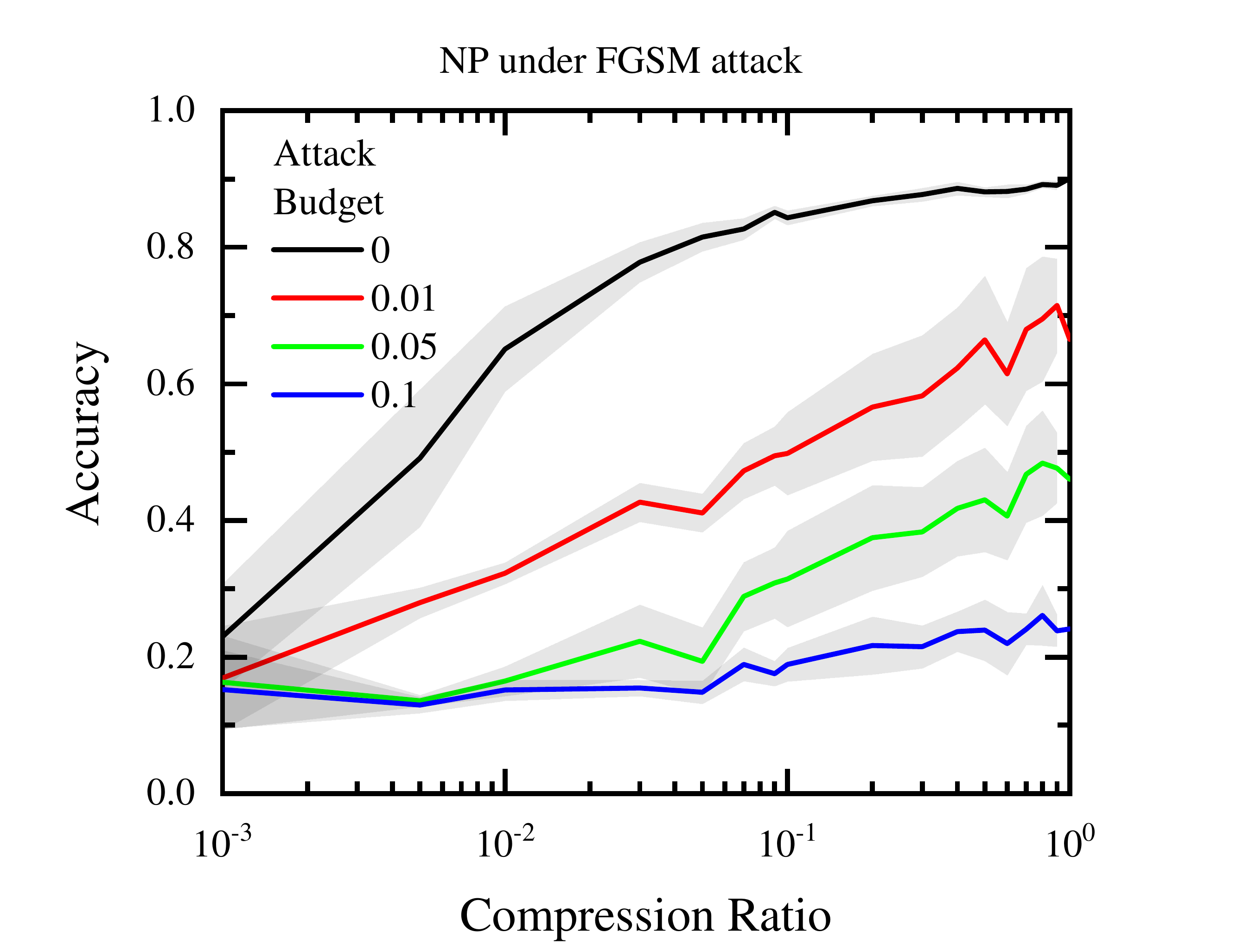}
  \includegraphics[width=0.3\linewidth]{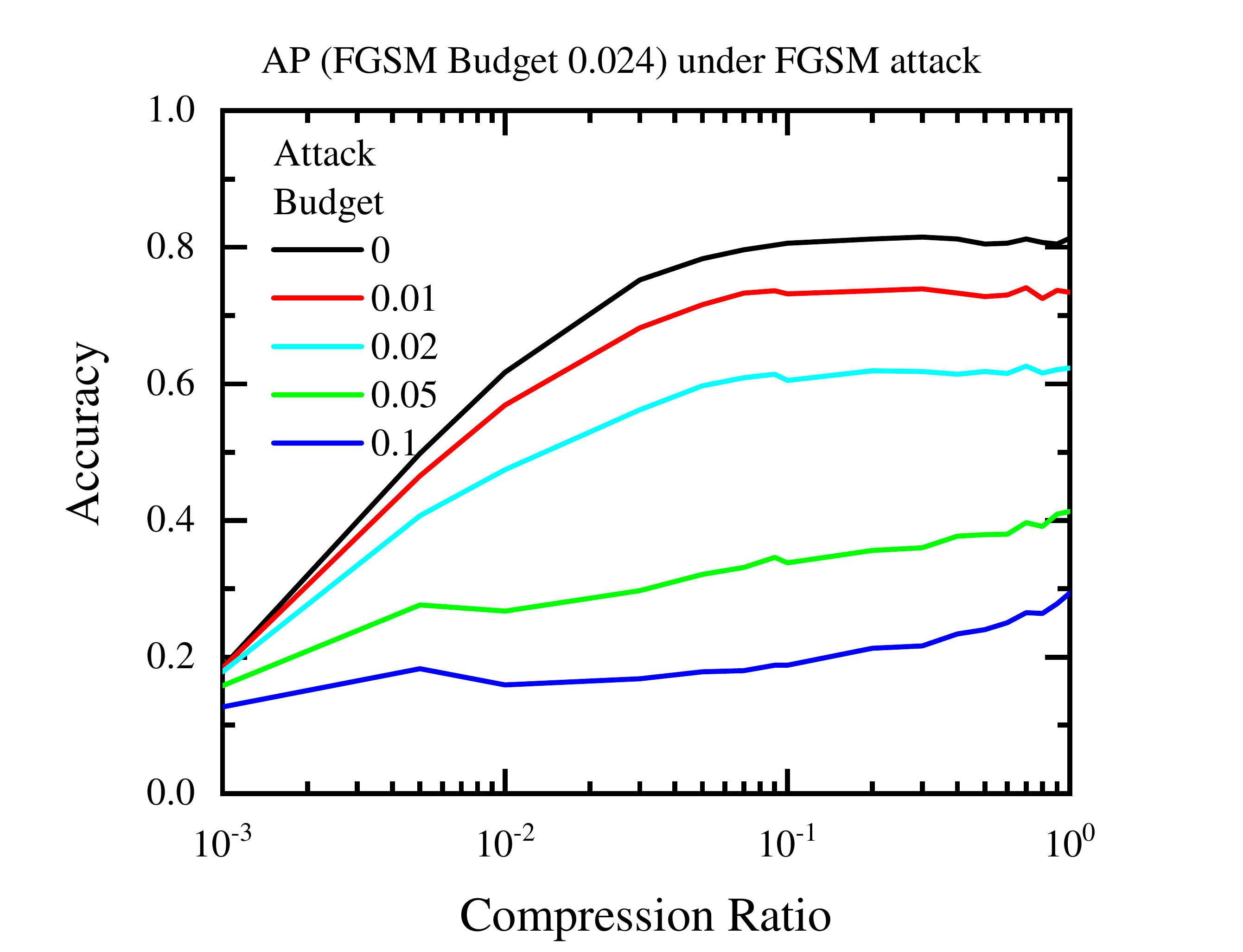}
  \includegraphics[width=0.3\linewidth]{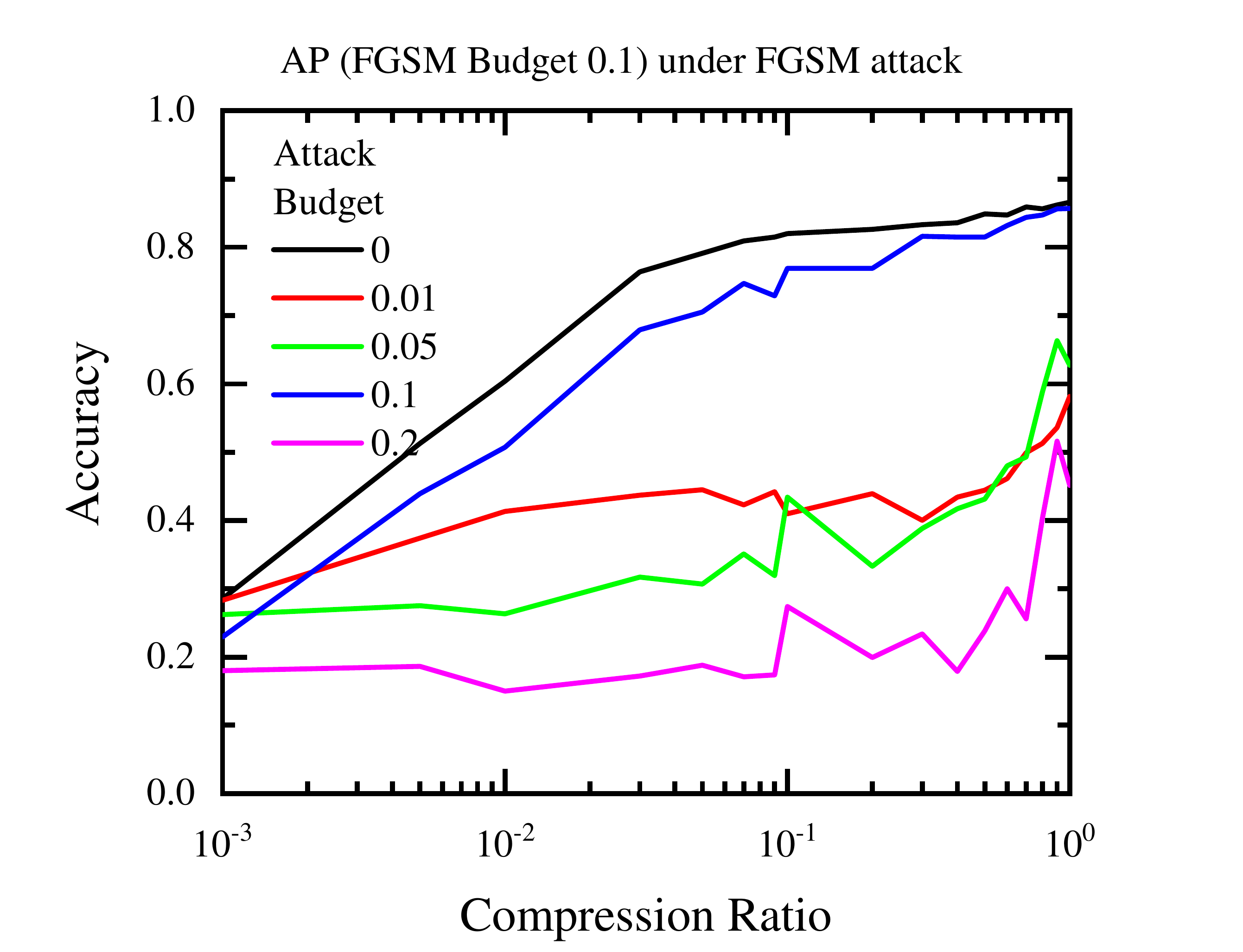}
  \includegraphics[width=0.3\linewidth]{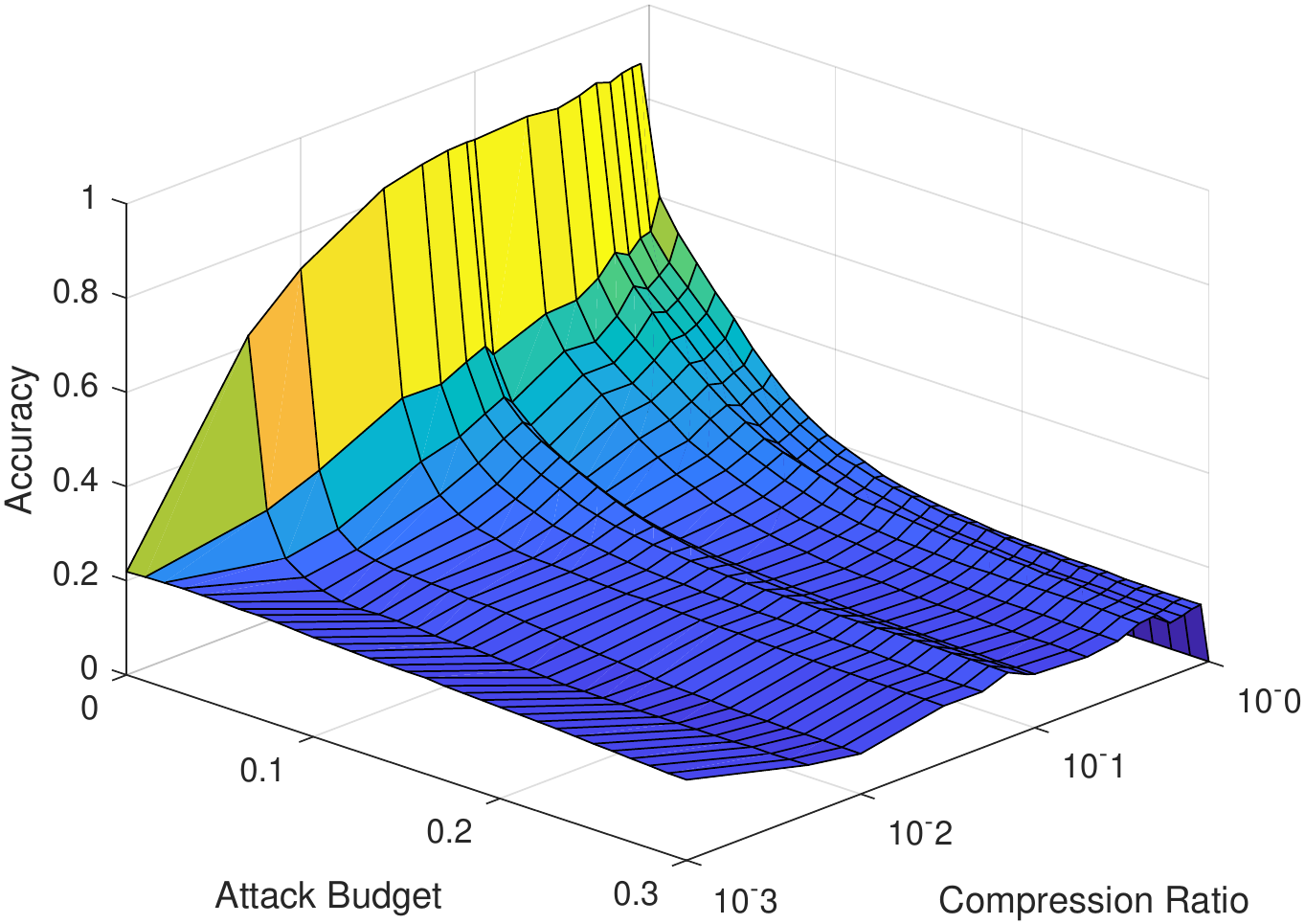}
  \includegraphics[width=0.3\linewidth]{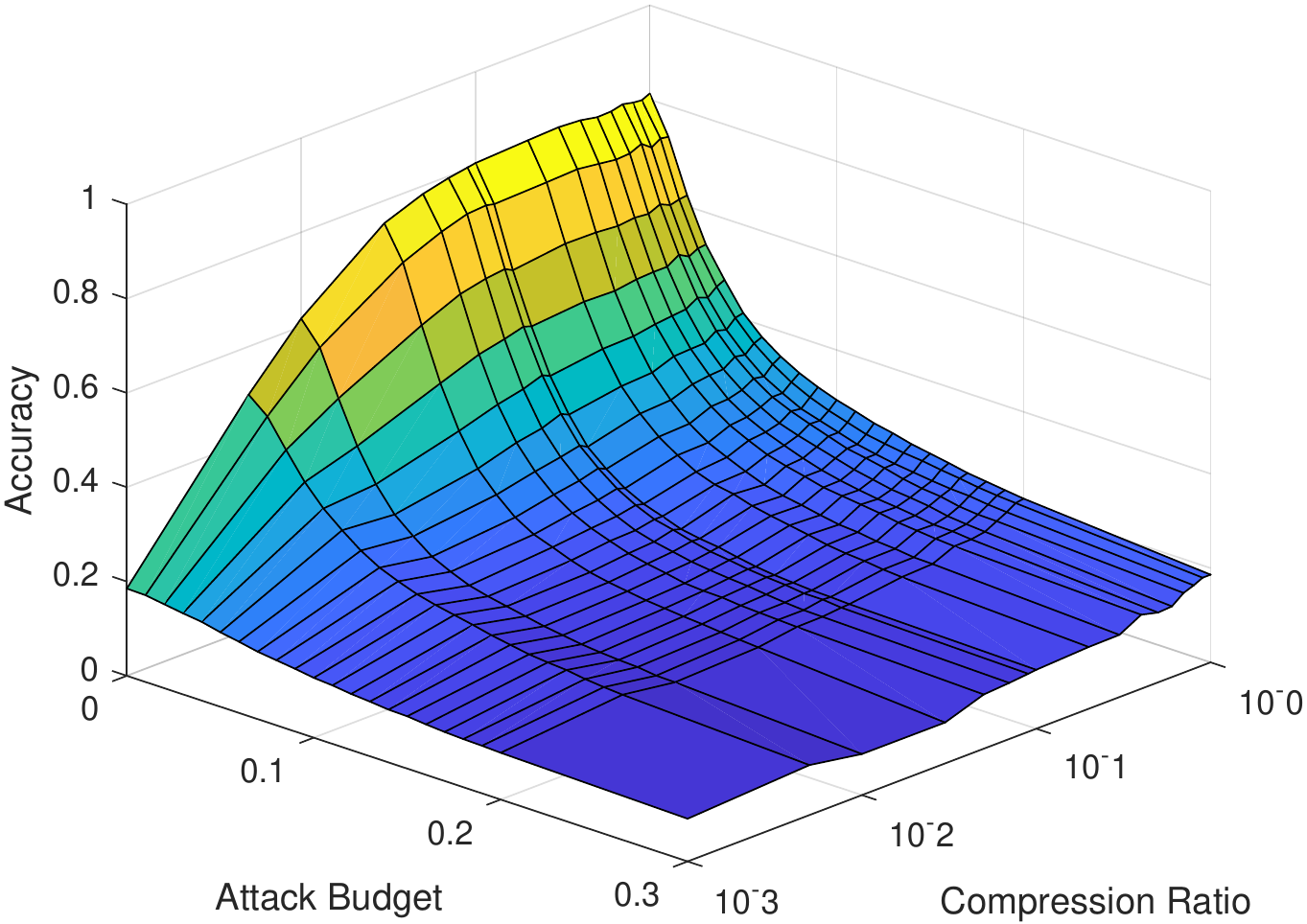}
  \includegraphics[width=0.3\linewidth]{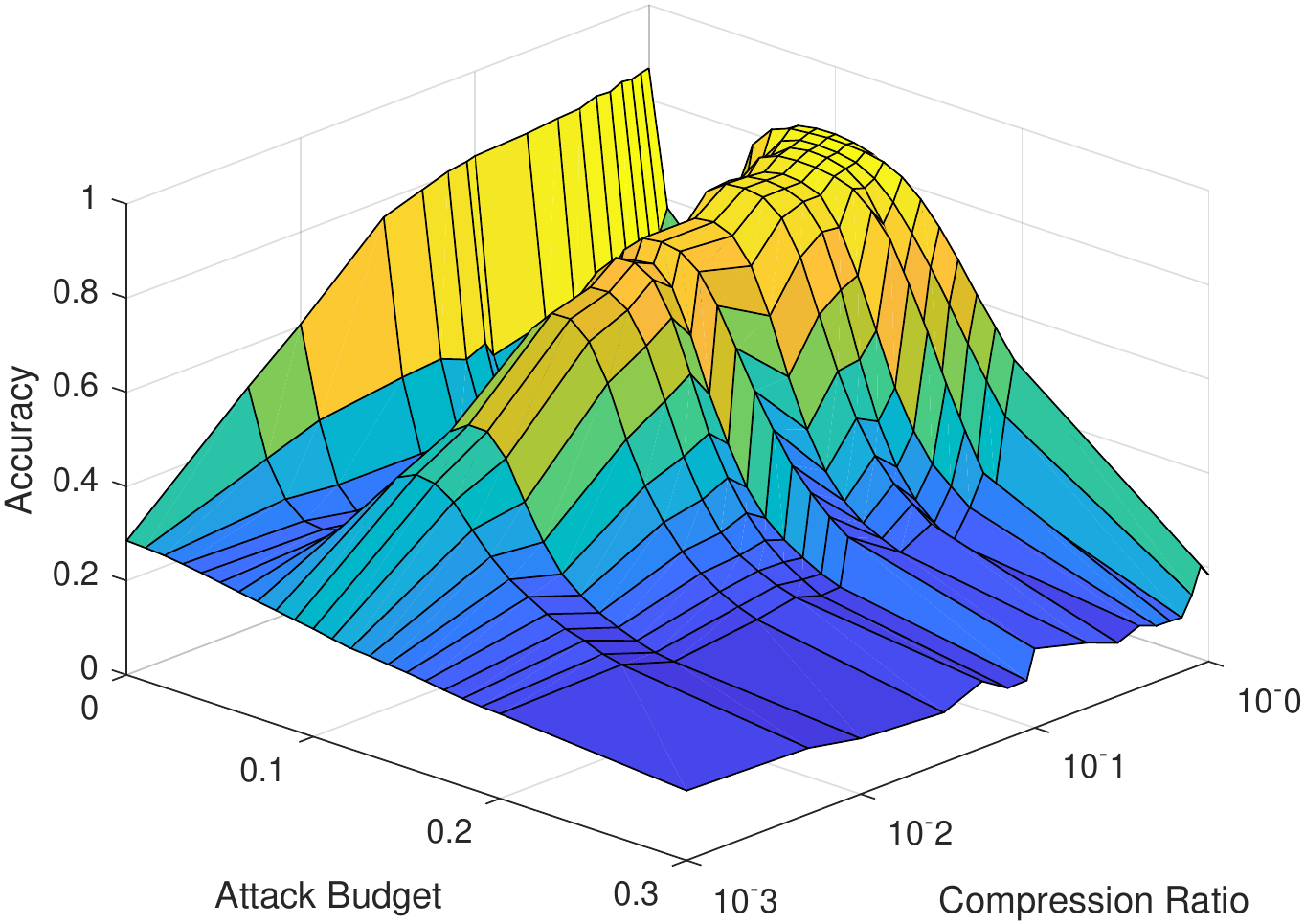}
  \caption{Comparison between the NP model and AP model (FGSM budgets $\varepsilon=0.024 \& 0.1$) of different compression ratios and different attack budgets for \textbf{CIFAR-10} datasets. 
  Each column corresponds to a different training method, while in each column: 
  (Upper) The solid colored lines and the corresponding gray area represent the average and standard deviation of the accuracy with different attack budgets obtained by independently repeating the pruning process 5 times for NP and 1 time for AP. 
  The wider the gray area is, the more unstable the model’s robustness is.
  (Lower) Three-dimensional representation of AER.
  }
  \label{fig:AT-FGSM-C10}
\end{figure*}

\begin{figure*}[t]
  \centering
  \includegraphics[width=0.3\linewidth]{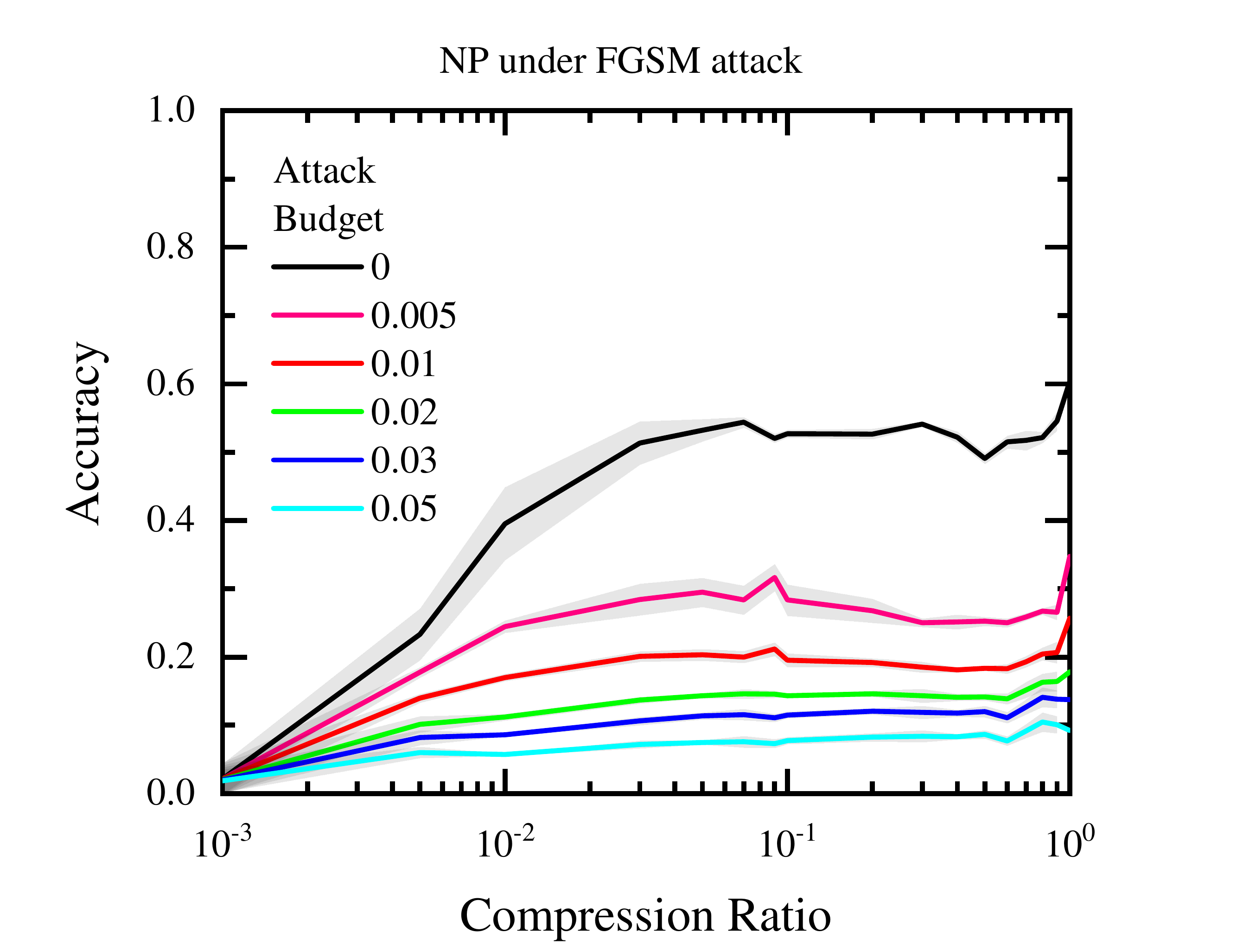}
  \includegraphics[width=0.3\linewidth]{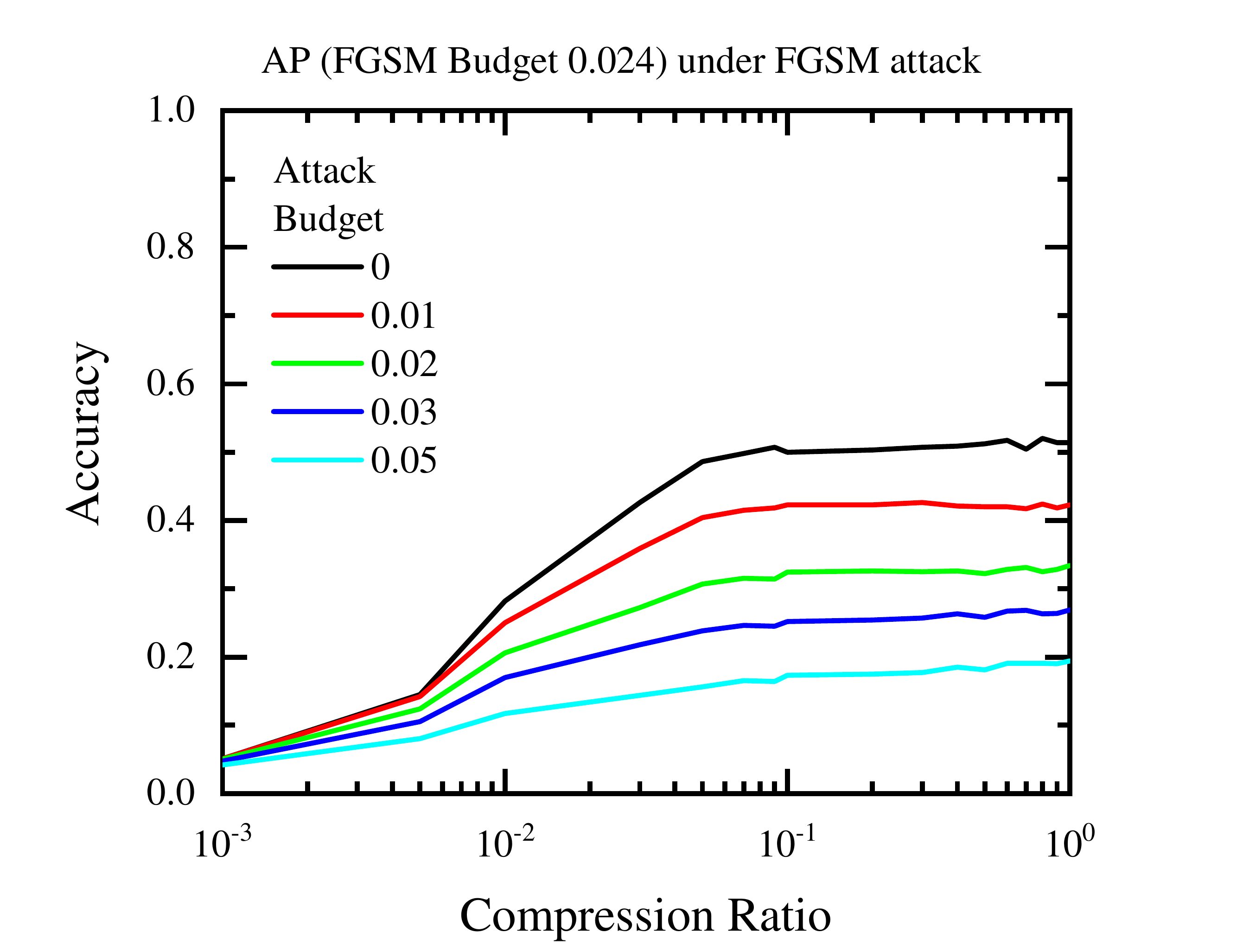}
  \includegraphics[width=0.3\linewidth]{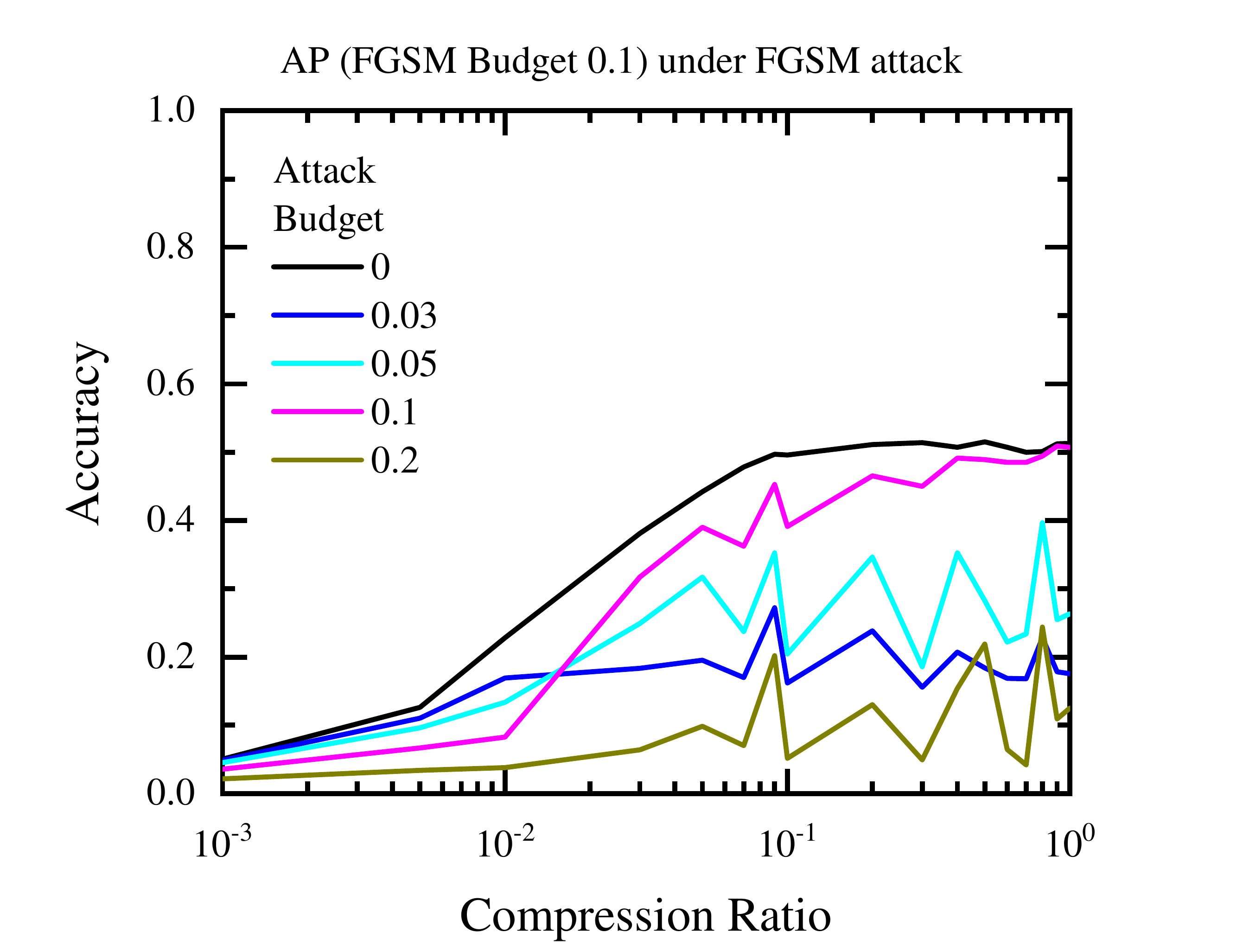}
  \includegraphics[width=0.3\linewidth]{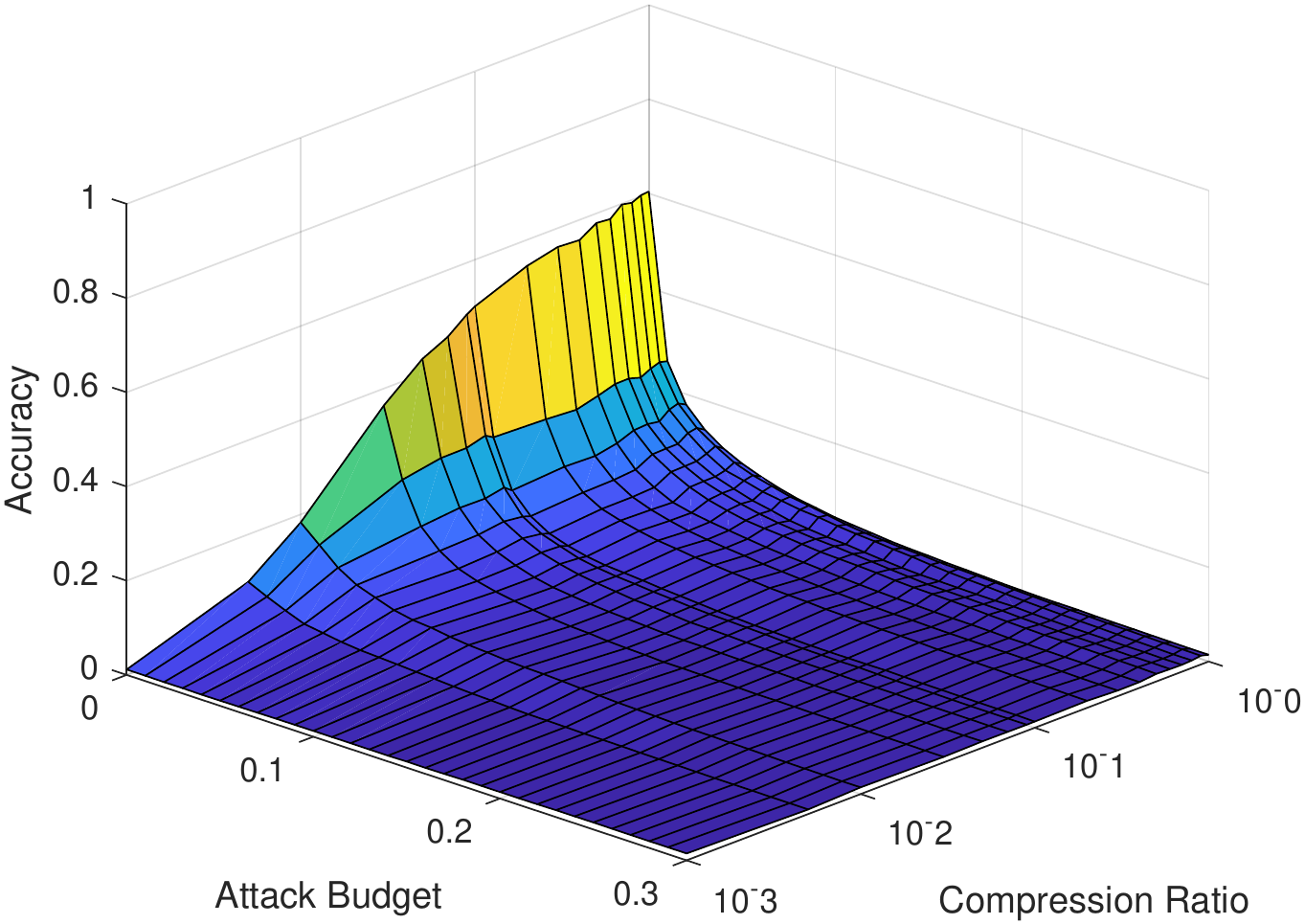}
  \includegraphics[width=0.3\linewidth]{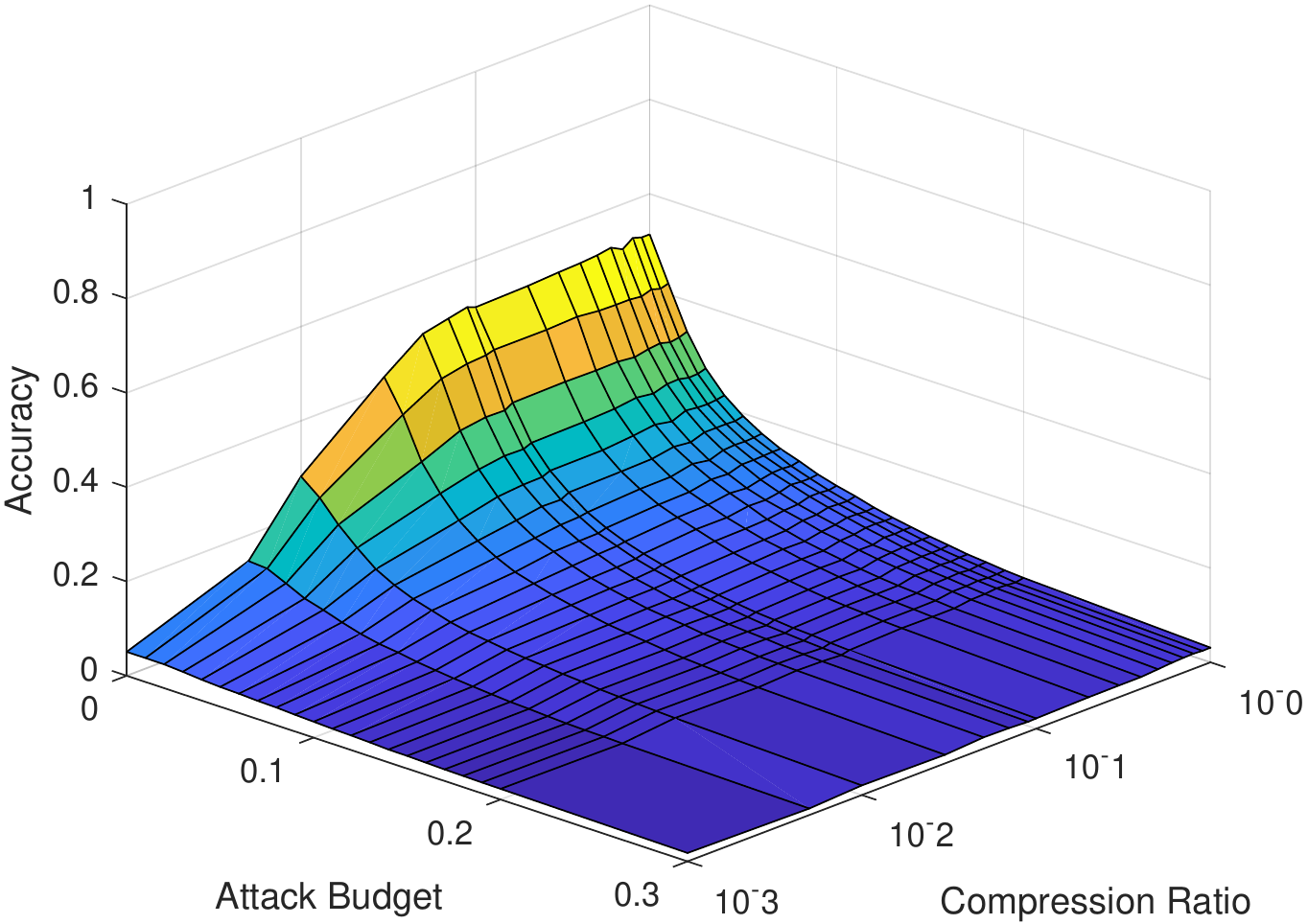}
  \includegraphics[width=0.3\linewidth]{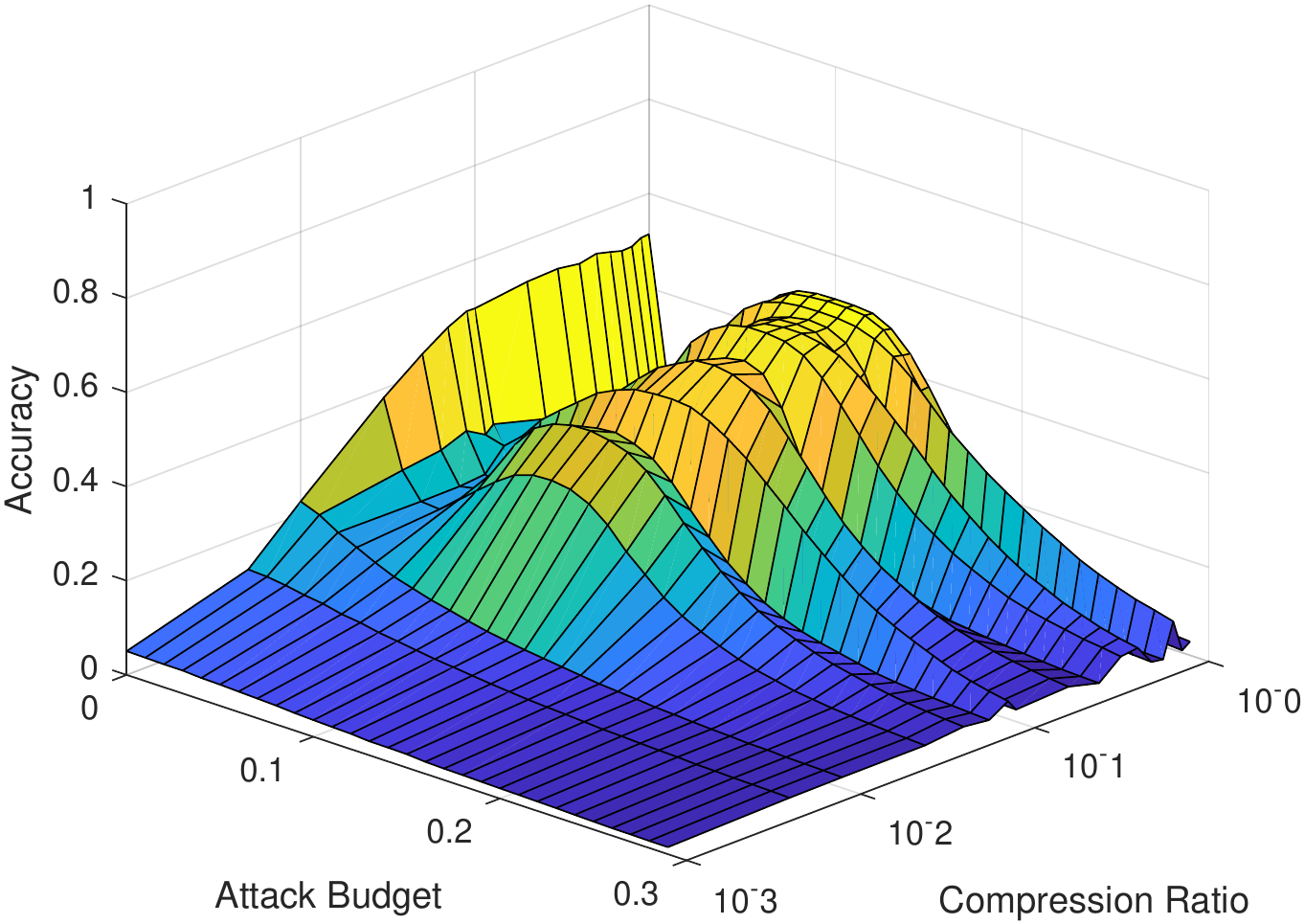}
  \caption{Comparison between the NP model and AP model (FGSM budgets $\varepsilon=0.024 \& 0.1$) of different compression ratios and different attack budgets for \textbf{CIFAR-100} datasets. 
  Each column corresponds to a different training method, while in each column: 
  (Upper) The solid colored lines and the corresponding gray area represent the average and standard deviation of the accuracy with different attack budgets obtained by independently repeating the pruning process 5 times for NP and 1 time for AP. 
  The wider the gray area is, the more unstable the model’s robustness is.
  (Lower) Three-dimensional representation of AER.
  }
  \label{fig:AT-FGSM-C100}
\end{figure*}

First, we analyze NP and AP approaches facing FGSM attacks on MNIST, CIFAR-10 and CIFAR-100 datasets as in Figure~\ref{fig:AT-FGSM}-\ref{fig:AT-FGSM-C100} to show the effects of AER, and we display the results about accuracy, efficiency and robustness to analyze the relationships among them. To show the stability of the algorithm more clearly, we repeat each method several times, using the solid line to represent the average results and the shaded area to represent the standard deviation of these results.
At the same time, we show the results of AER for three-dimensional surfaces, which can more clearly display the trend of the accuracy as the model compression ratio and attack budget change.

For each subfigure, different attack budgets have different accuracy-compression ratio curves. NP only uses clean data in training, while AP (FGSM) uses both clean data and prescribed budget AEs in training, and thus, the accuracy of the clean data and trained budget AEs is the highest, with minor fluctuations. 
In contrast, accuracy under other attack budgets is lower, with large fluctuations. 
This result is due to the training process only considering clean data and AEs with the budget; the pruning process disregards the weights that have little impact on these data, but these lost weights may be very important for attacks with other budgets. 
It can be concluded that the redundant weights in loss do not indicate unimportance for attack. 
Neither NP nor AP methods can ensure that the pruned model has good robustness, and the results \textbf{drastically vary} (exhibit great fluctuation or instability) with different pruning processes (training a random initial model to one with high accuracy and then gradually pruning it to one with the target compression ratio), especially under attacks of large strength.

By comparing the results of these datasets and carefully observing each accuracy-compression ratio curve, we can divide the curve into the stages \textbf{stable, falling and failing}, corresponding to no significant changes in accuracy, rapid declines and complete failure of the model, respectively.
Curves of different budgets correspond to different inflection points in the stable and falling stages. 
Both NP and AP perform best with respect to the trained clean data and AEs with the budget, exhibiting the longest stable stages and inflection points which are more to the left. AP can only significantly improve the robustness of the budget. 
For curves of different budgets, the farther the budget is from the prescribed budget or clean data, the faster it will enter  the falling stage. This effect was rarely mentioned in previous work; for example, Guo et al.~\cite{YiwenSparse} report that under FGSM attacks with $\ell_\infty$ norm $\varepsilon=0.1$, the pruned model with less than $2.5\%$ nonzero weights can achieve the best robustness. However, it can be found from the figure that at this time, the model only exhibits high robustness under the $0.1$ budget attack and that the robustness against other budgets has been greatly reduced.
Within the falling stage, the curves corresponding to all budgets decreased and approached.
Finally, in the failure stage, curves with different budgets are no longer significantly different, and they all decline quickly. This result is possibly due to gradient masking with such high-intensity compression ratios, resulting in FGSM method failure; at the same time, the model at this stage has no practical application value. 

In summary, compared with the existing evaluation criteria discussed before, we can clearly observe that on the one hand, our evaluation criterion can more clearly reflect the comprehensive performance of the model in all aspects and comprehensively evaluate the accuracy performance of the model under different compression ratios and facing different budget attacks; on the other hand, the existing AP (FGSM) methods are based on the prescribed budget, and they can obviously only guarantee high clean accuracy and prescribed budget robustness but cannot ensure the robustness with respect to other budget attacks at all, which \textbf{indicates high sensitivity to the budget}.

\subsection{AER of NP, AP and BAP under DeepFool attack}

\begin{figure*}[!t]
  \centering
  \includegraphics[width=0.3\linewidth]{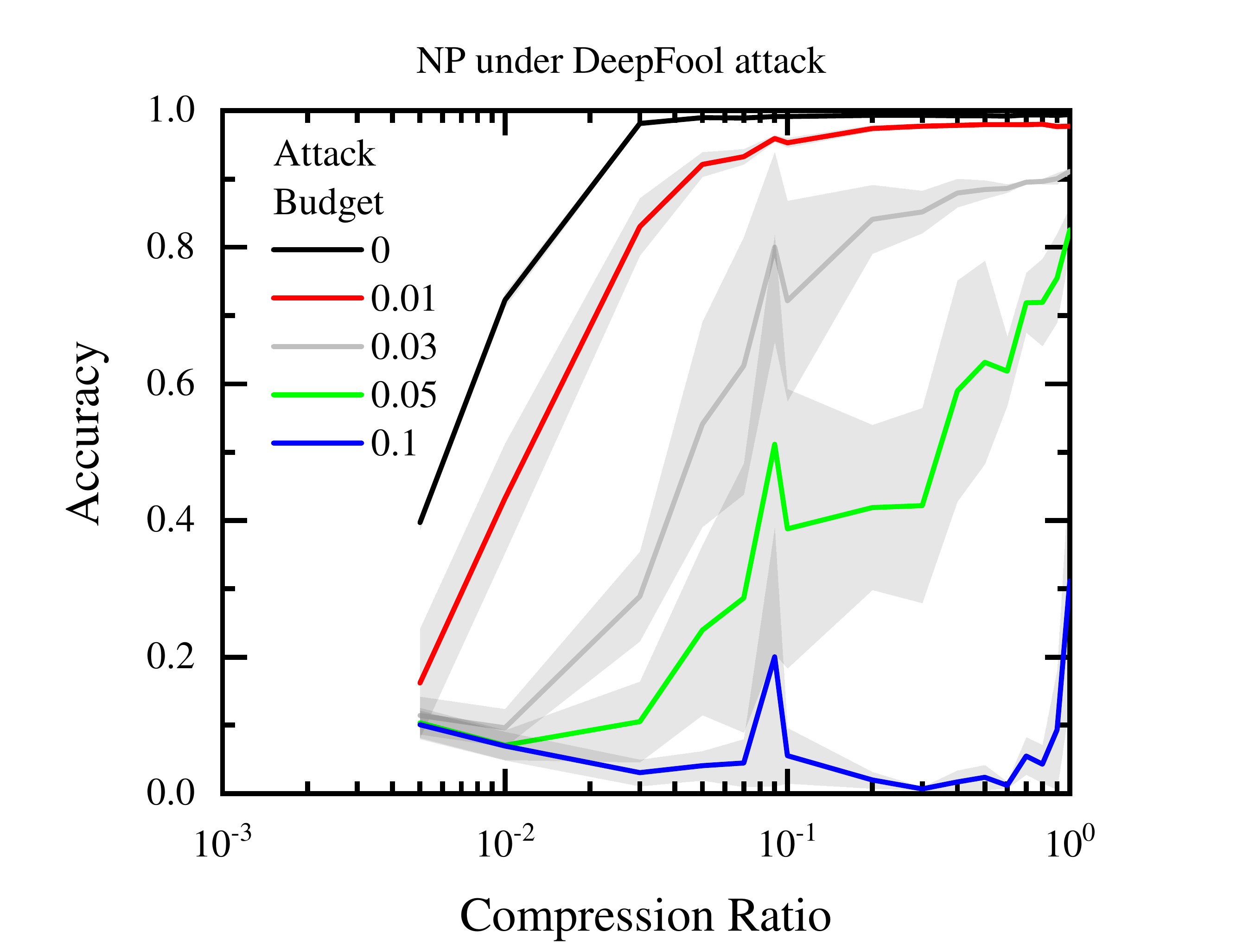}
  \includegraphics[width=0.3\linewidth]{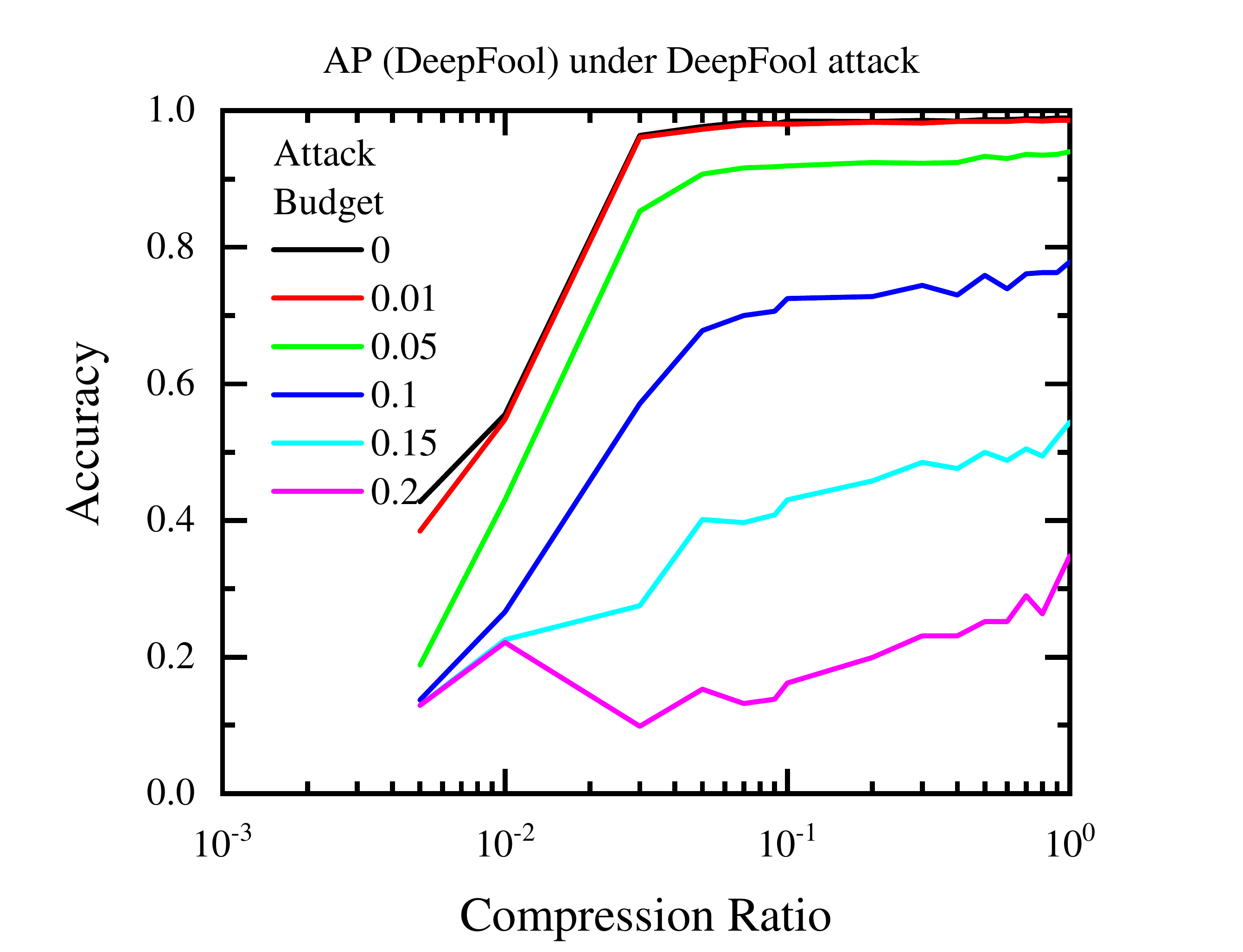}
  \includegraphics[width=0.3\linewidth]{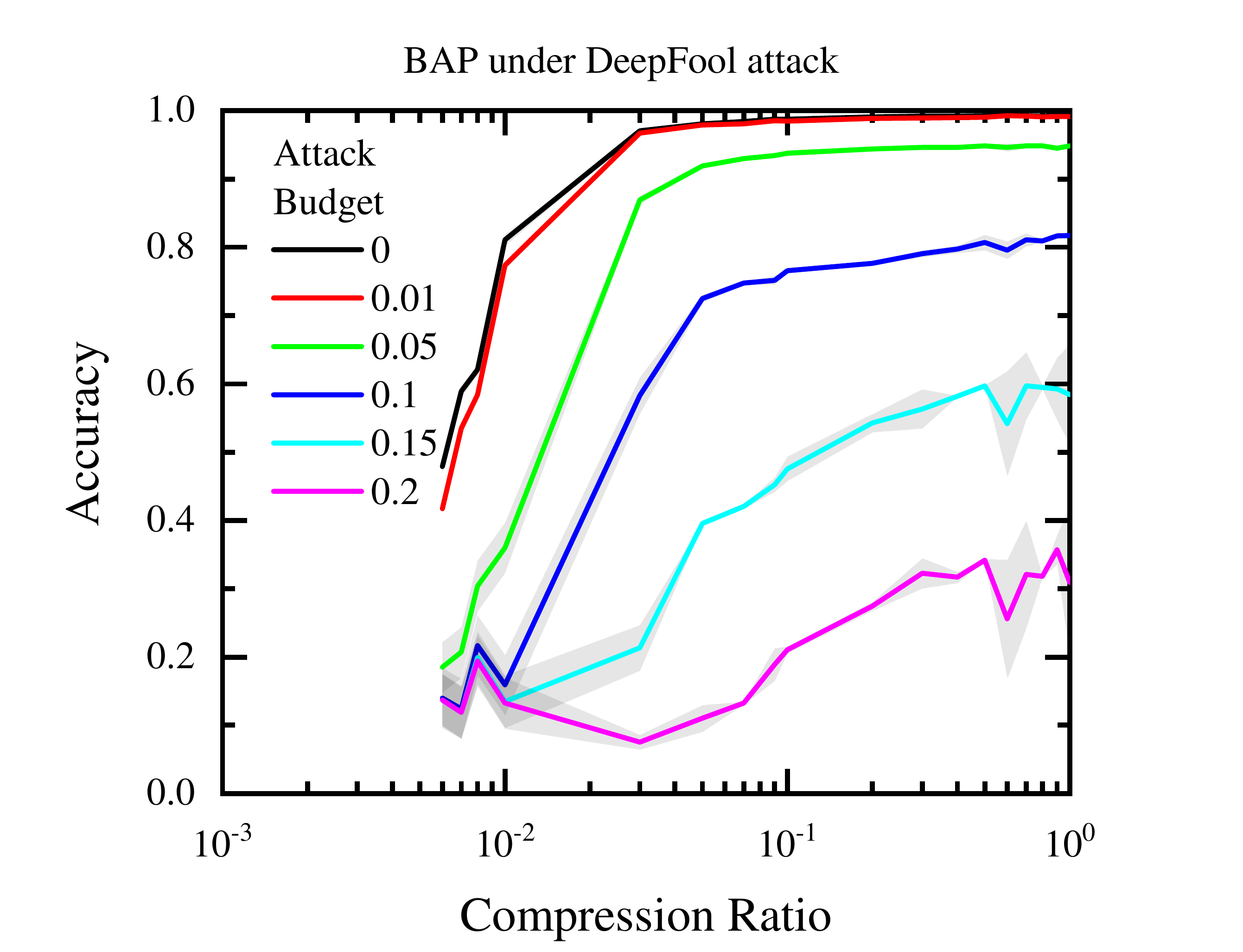}\\
  \includegraphics[width=0.3\linewidth]{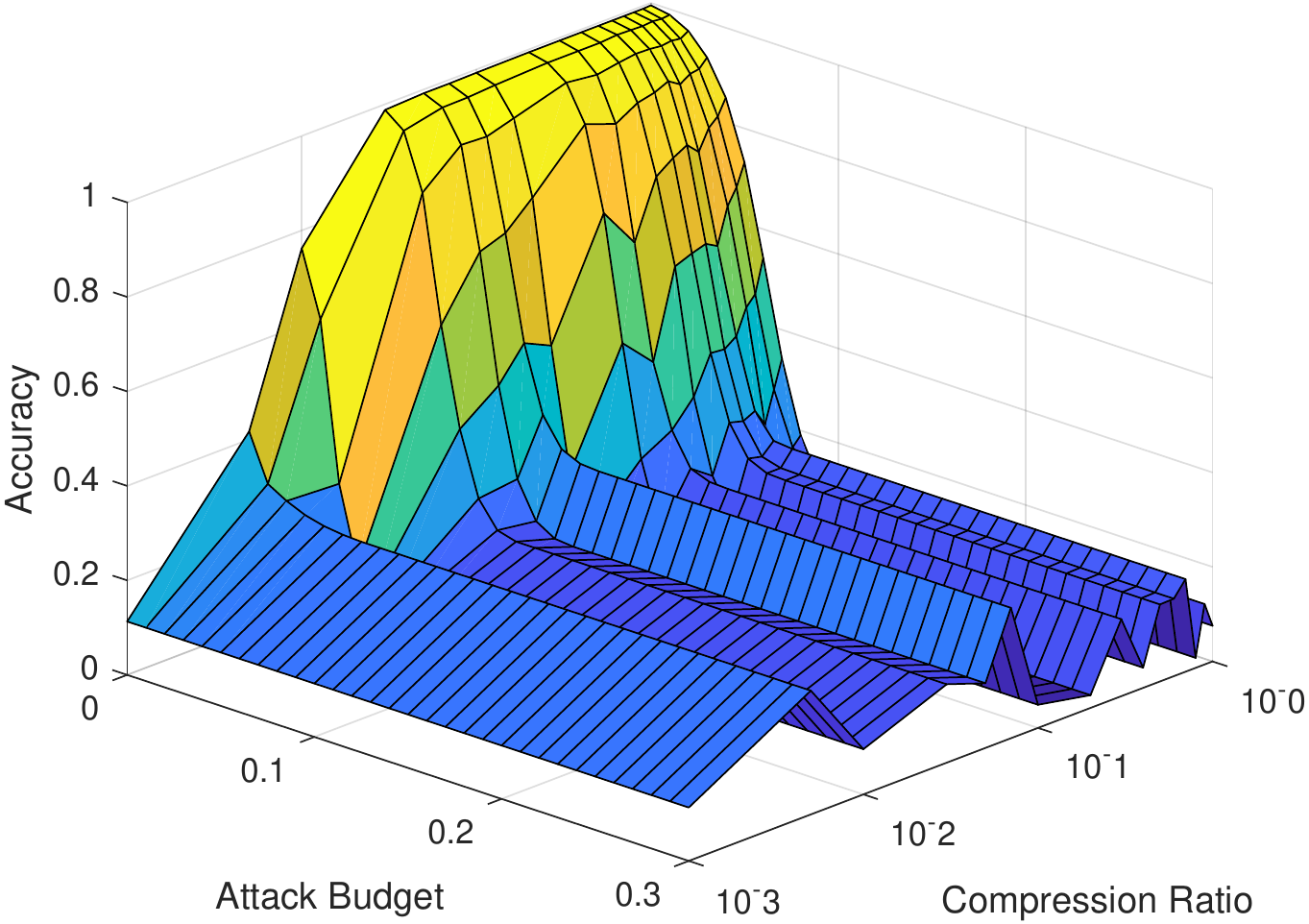}
  \includegraphics[width=0.3\linewidth]{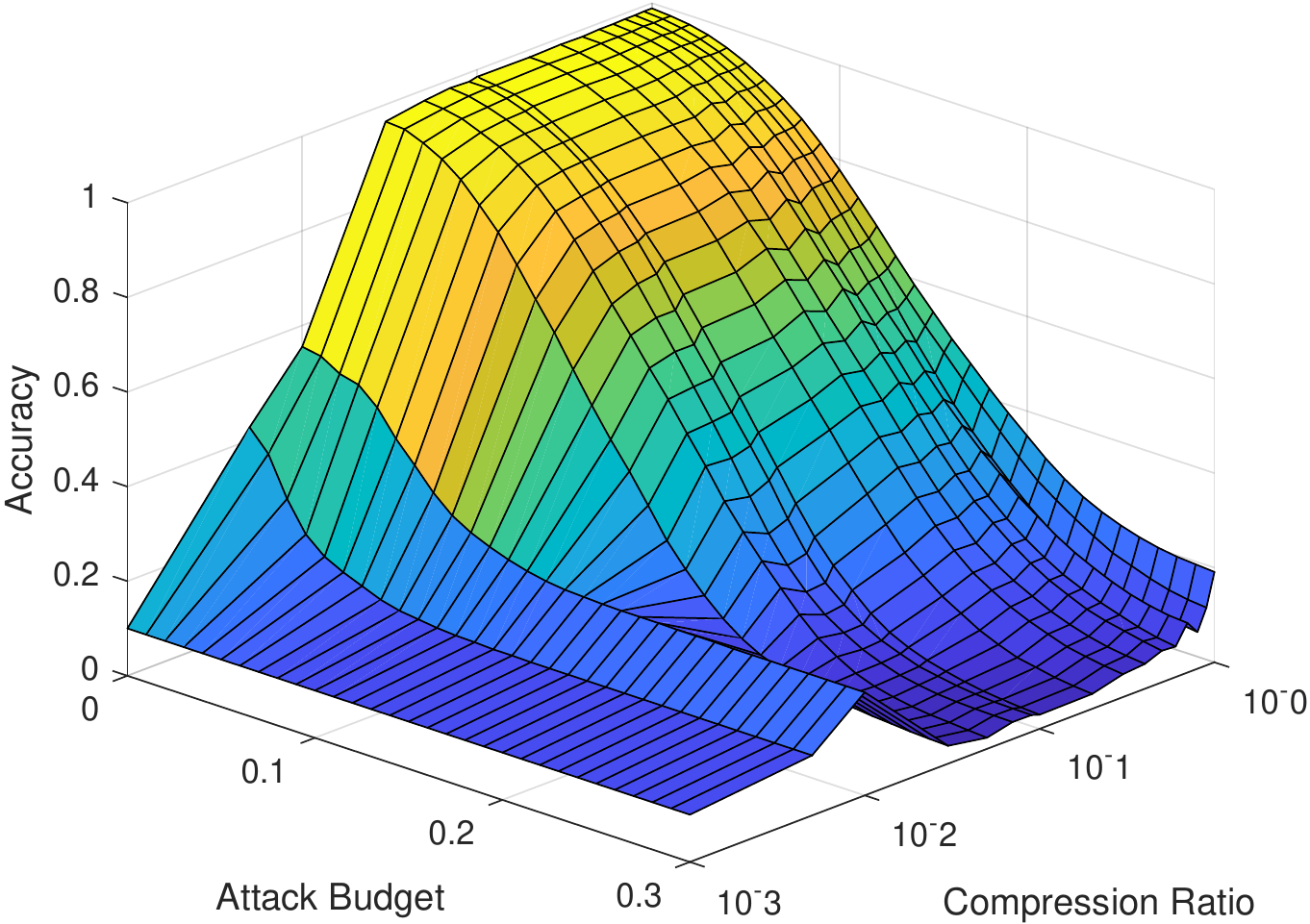}
  \includegraphics[width=0.3\linewidth]{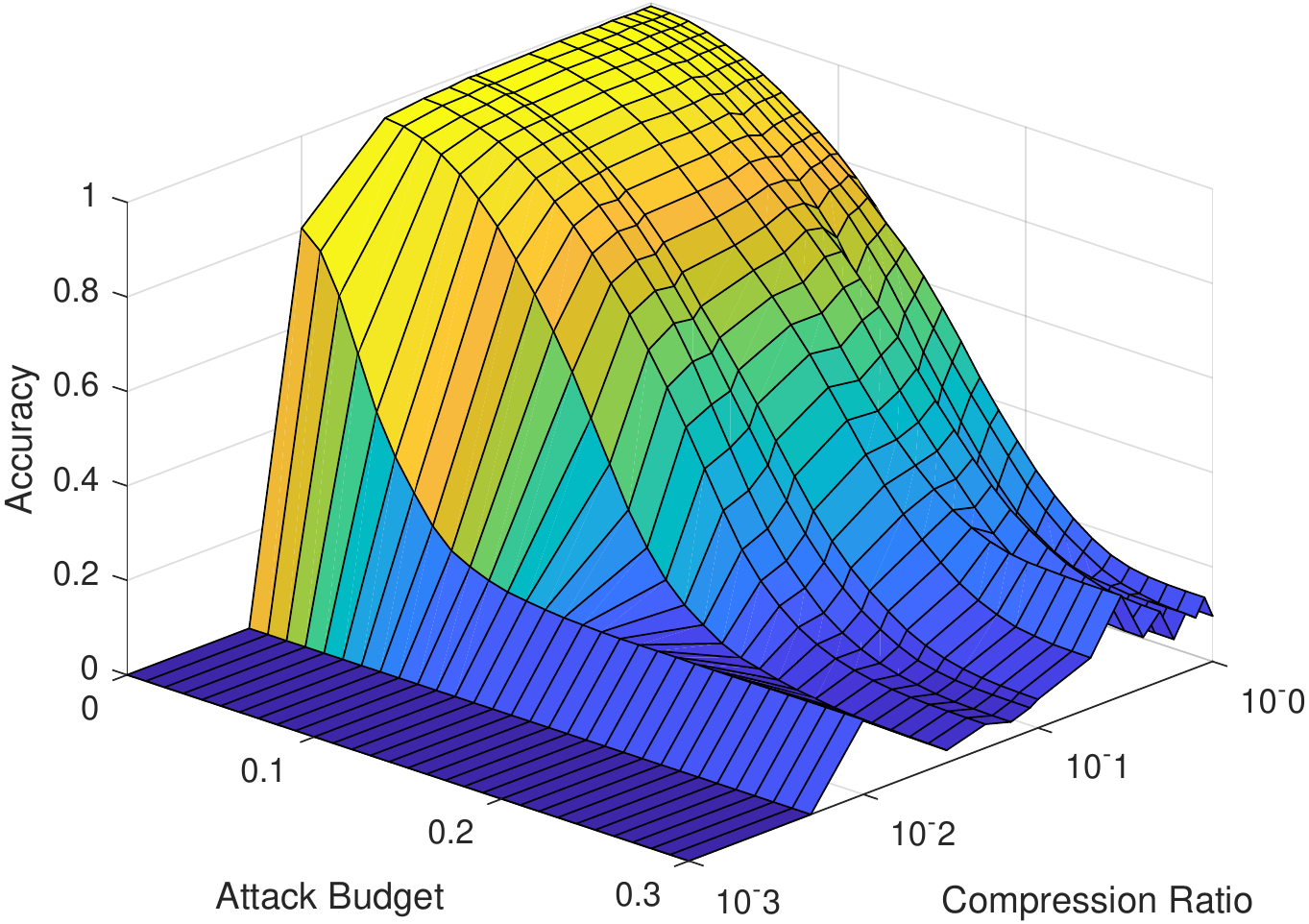}
  \caption{Comparison between BAP model, NP and AP (using DeepFool AEs) methods of different compression ratios and different attack budgets for \textbf{MNIST} datasets. 
  Each column corresponds to a different training method, while in each column: 
  (Upper)
  The solid lines in color and the corresponding gray area represent the average and standard deviation of accuracy with different attack budgets obtained by independently repeating the pruning process 5 times for NP, 1 time for AP and 2 times for BAP. 
  The wider the gray area is, the more unstable the model’s robustness.
  (Lower) Three-dimensional representation of AER. 
  }
  \label{fig:BAT}
\end{figure*}

\begin{figure*}[!t]
  \centering
  \includegraphics[width=0.24\linewidth]{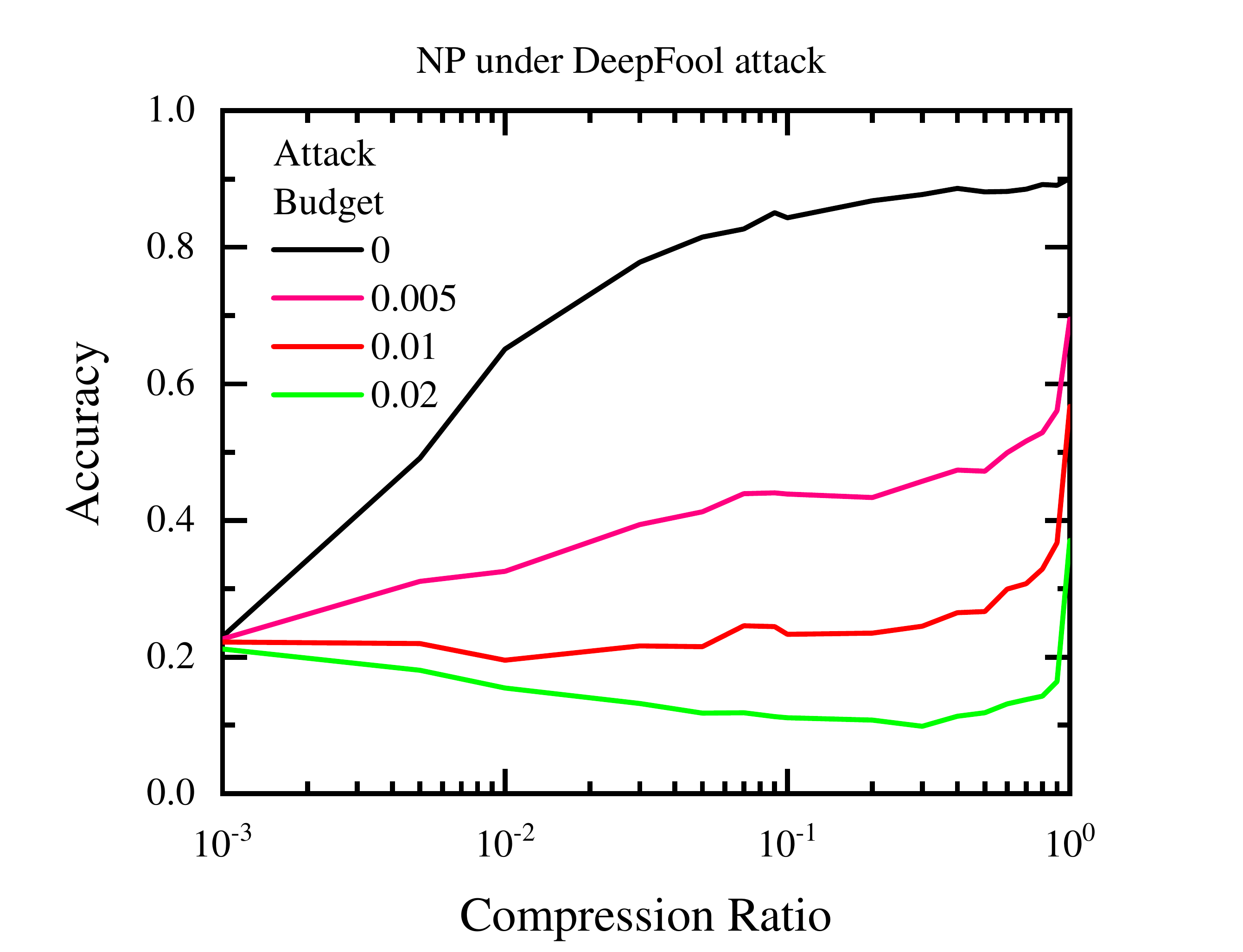}
  \includegraphics[width=0.24\linewidth]{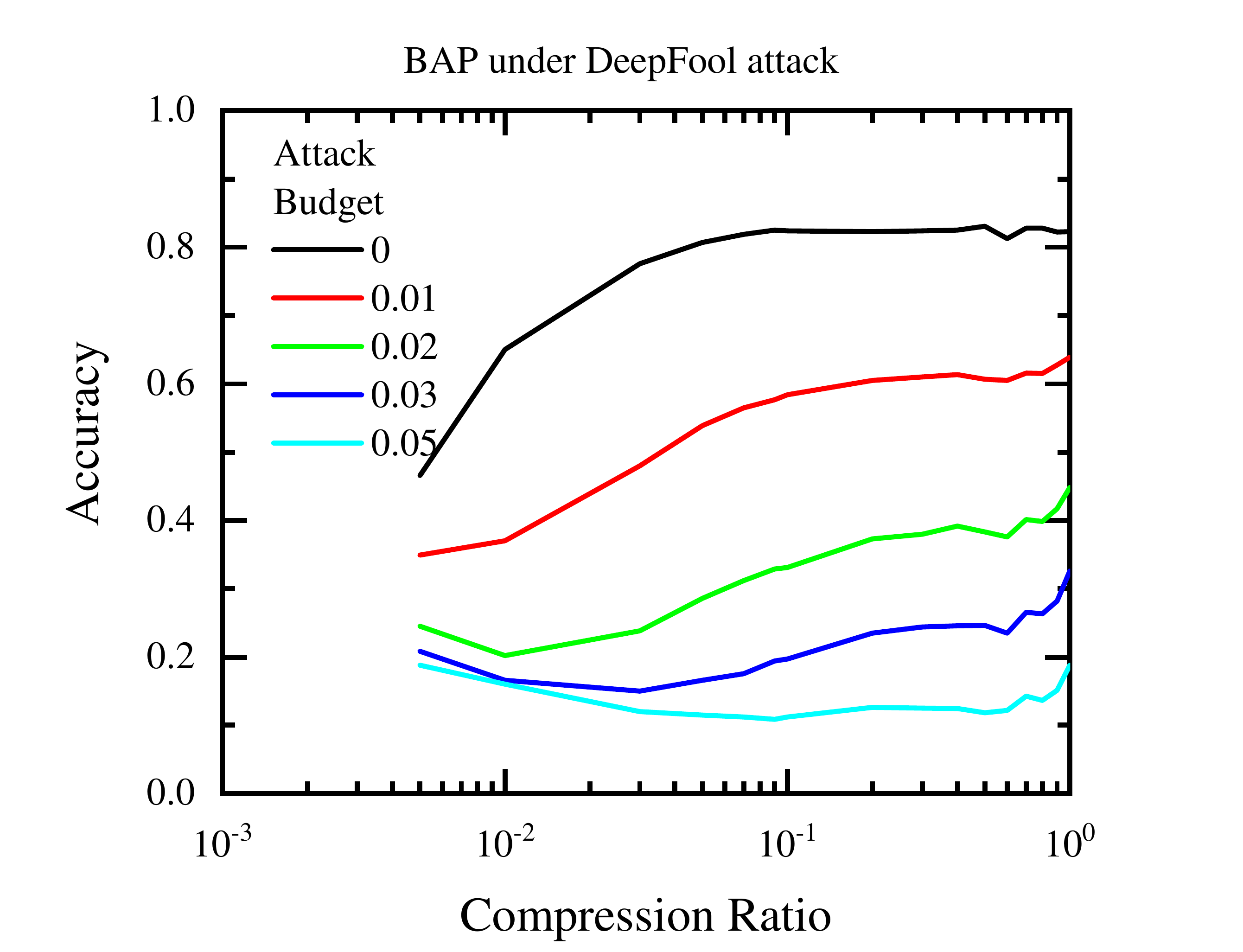}
  \includegraphics[width=0.24\linewidth]{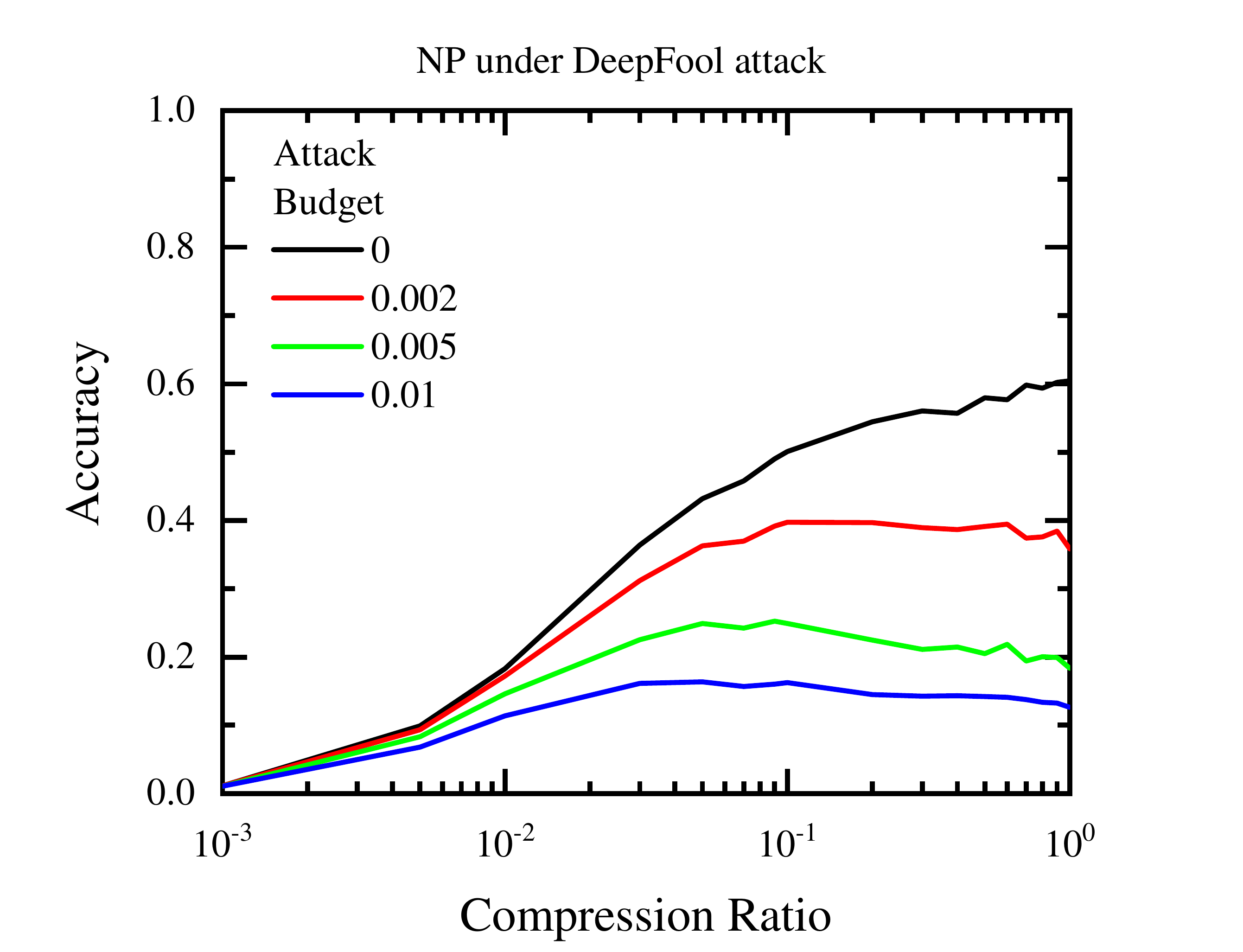}
  \includegraphics[width=0.24\linewidth]{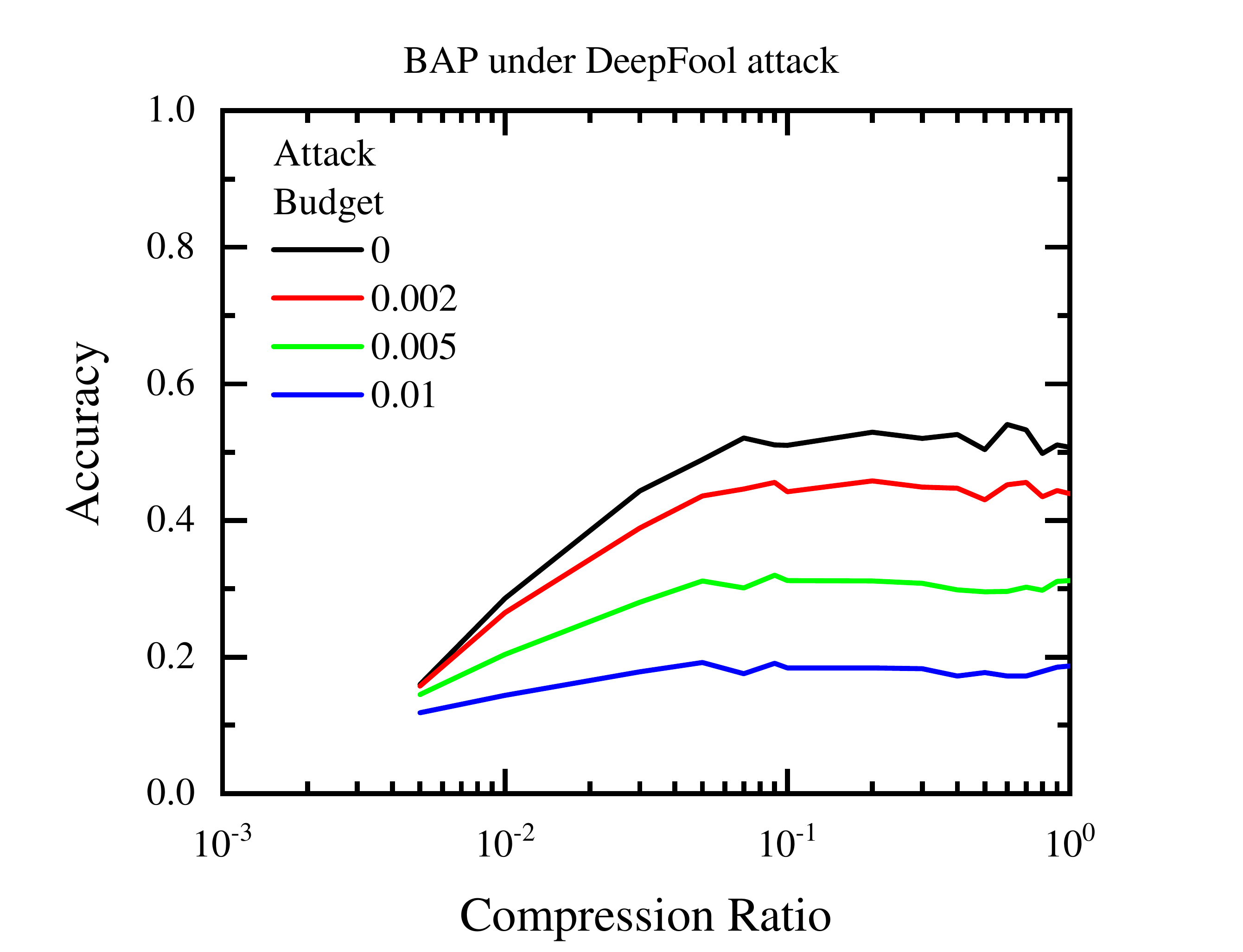}\\
  \includegraphics[width=0.24\linewidth]{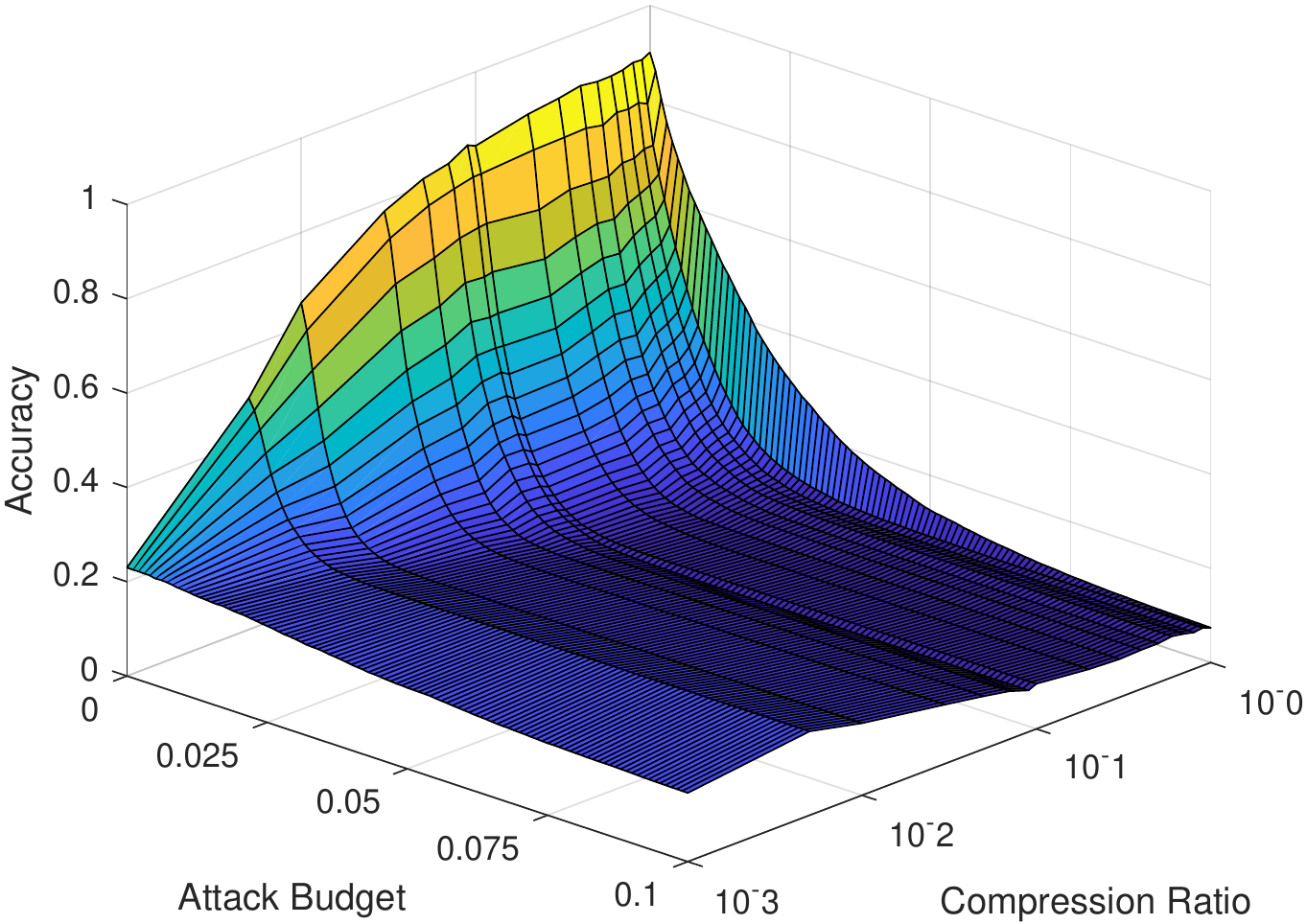}
  \includegraphics[width=0.24\linewidth]{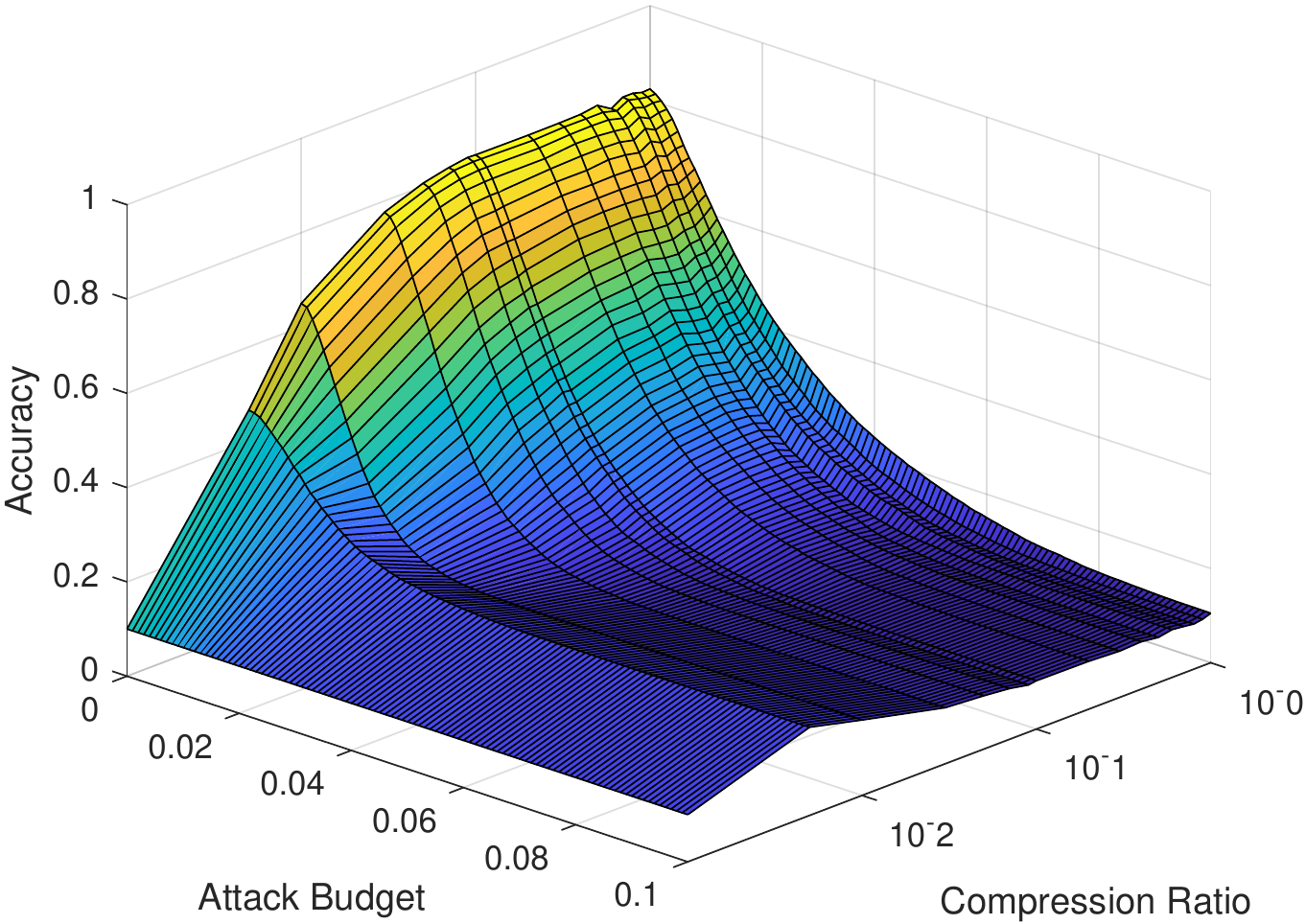}
  \includegraphics[width=0.24\linewidth]{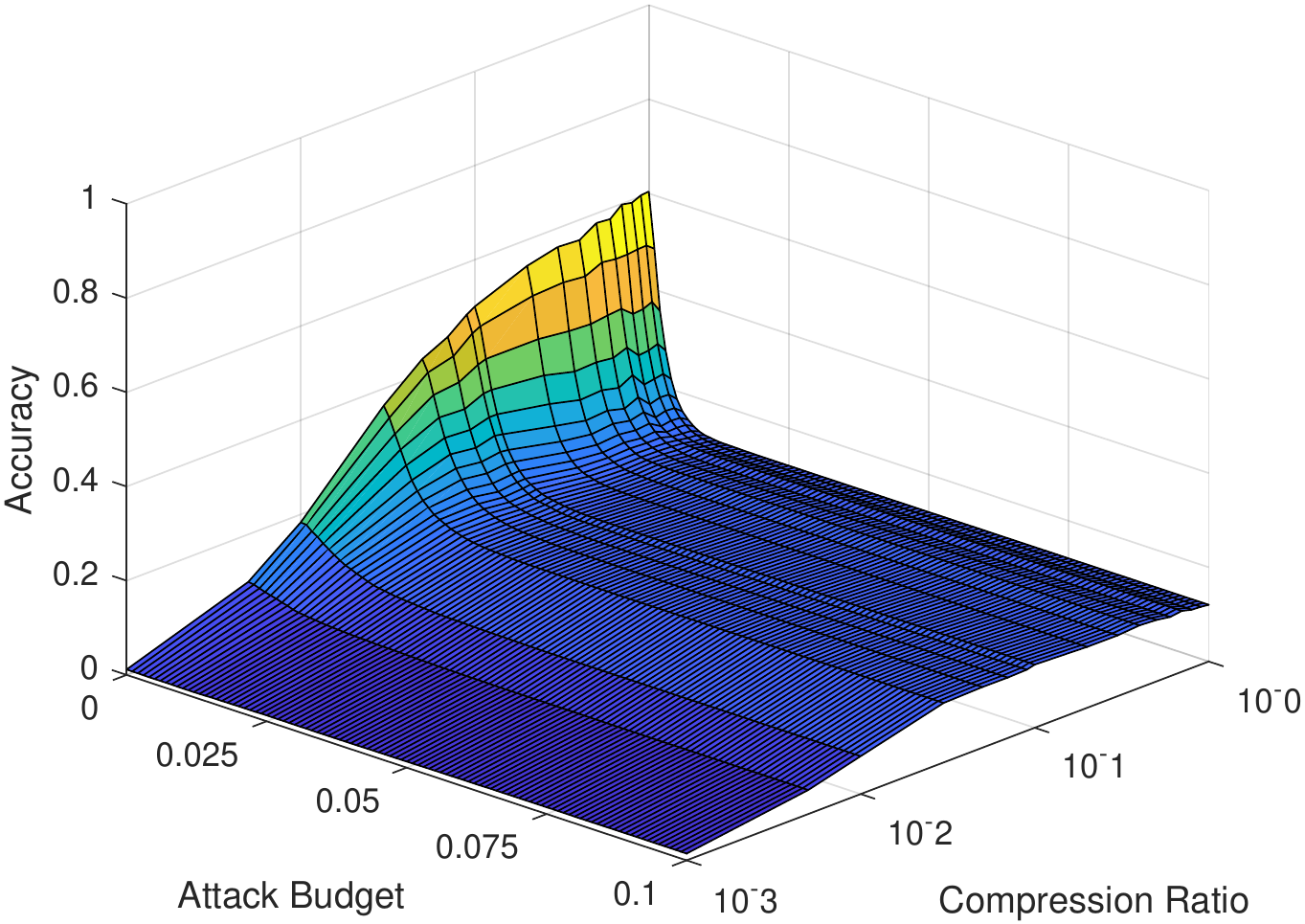}
  \includegraphics[width=0.24\linewidth]{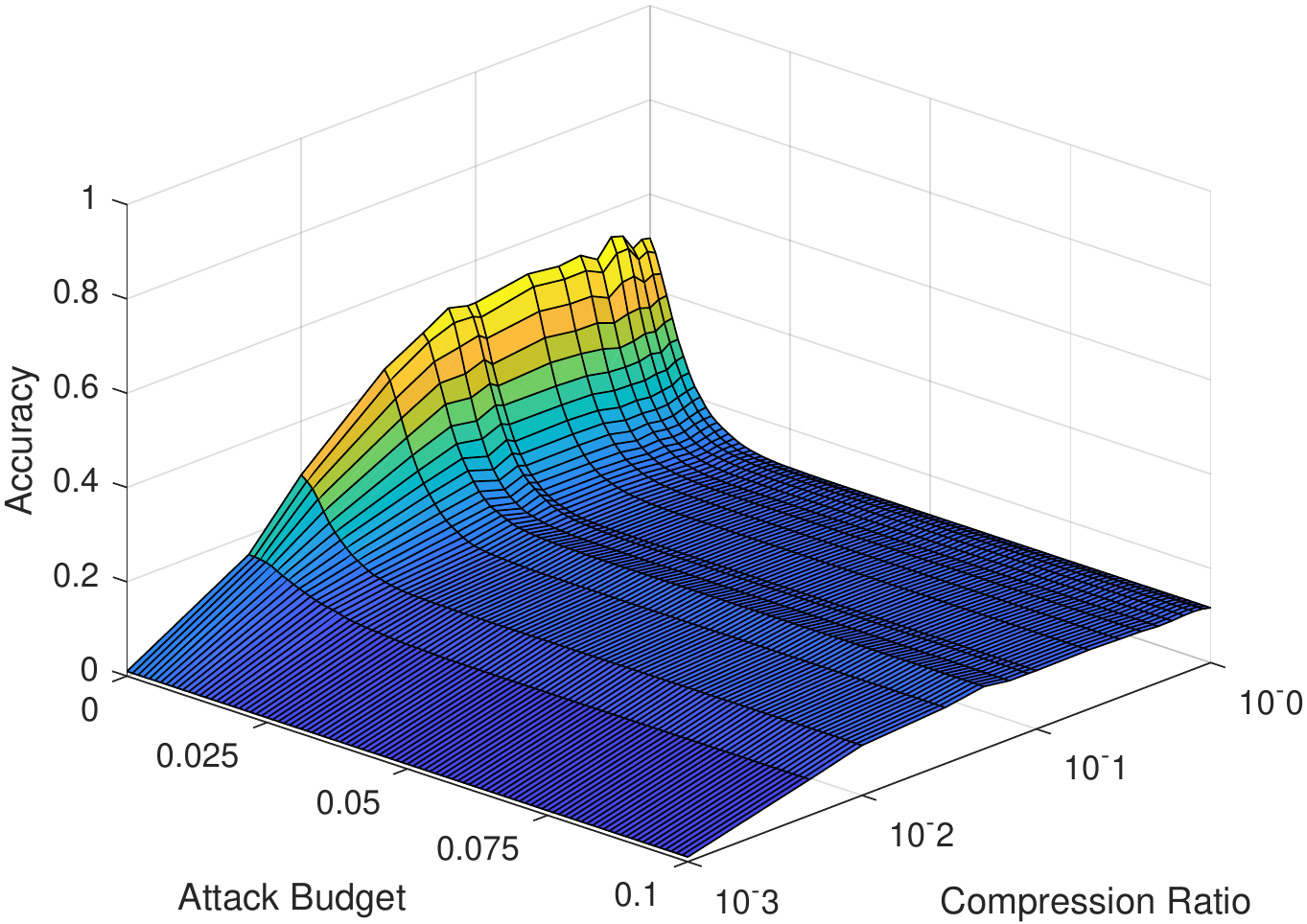}
  \caption{Comparison of BAP model with NP of different compression ratios and different attack budgets for \textbf{CIFAR-10 and CIFAR-100} datasets. 
  Each column corresponds to a different dataset with a different training method, while in each column:
  (Upper) The solid lines in color correspond to the accuracy with different attack budgets, as in the legend. 
  (Lower) Three-dimensional representation of AER.
  }
  \label{fig:BAT2}
\end{figure*}

Next, we show the BAP results in Figure~\ref{fig:BAT} \& \ref{fig:BAT2}. 
Overall, from the three-dimensional figures, under the same compression ratio and attack budget, BAP achieves the highest accuracy in most cases with less fluctuations. 
This result is expressed as the overall AER surface moving away from the origin. 
In a sentence, it is shown that the model obtained by our BAP exhibits better comprehensive AER performance. 

The most notable feature of the result of BAP compared with the AP methods is that there is no prescribed budget, and the model's robustness to different attack budgets is improved by adaptively estimating a nonuniform budget.
Judging from the results, BAP will not produce two peaks corresponding to budgets like the AP (FGSM) method, as shown in Figure~\ref{fig:AT-FGSM-C10}~\&~\ref{fig:AT-FGSM-C100} with budget $0.1$. With an increase in the budget, the performance of different budget attacks is improved and the accuracy tends to monotonically decrease. 
By carefully observing the value of the results in Table~\ref{compare}, we can find that the clean accuracy of the results trained by BAP is significantly higher than that of AP (DeepFool), which indicates less loss of accuracy caused by the BAP AEs.
Overall, BAP can achieve clean accuracy very close to that of NP and at the same time can achieve accuracy close to that of AP (DeepFool) in the face of a very large budget attack. On this basis, compared with that of NP and AP (DeepFool), the overall accuracy performance of BAP with clean data and different budget attacks is very prominent, ranking first in most cases in Table~\ref{compare}, and BAP is obviously stronger than NP and AP (DeepFool).
BAP fails to perform best with respect to individual data, which will be the subject of further improvement in the future.

\begin{table*}[!t]
  \caption{
  Comparison of the $\text{AER}(\alpha, \varepsilon)$ and avg-AER$(\Gamma, \Theta)$ of BAP with AP (DeepFool) and NP for various datasets,
  various compression ratios and various $\ell_2$ white-box DeepFool attack budgets. 
  Note that $\text{AER}(\alpha, \varepsilon)$ with $\varepsilon = 0$ corresponds to clean accuracy and that nonzero $\varepsilon$ corresponds to accuracy of AEs with varying attack budgets.
  We chose $2$ different representative $\alpha$ and $\varepsilon$ combined with $\alpha=1$ and $\varepsilon=0$ to show the performance of $\text{AER}(\alpha, \varepsilon)$ and the corresponding avg-AER$(\Gamma, \Theta)$.
  We indicate the highest result with bold red and the second highest result with blue underline.
  }
\label{compare}
\vspace{0.5cm}
\centering
\setlength{\tabcolsep}{1.0mm}{
\begin{tabular}{|c|c|cc|ccc|ccc|ccc|}
\hline
& Defense     & \multicolumn{2}{c|}{avg-AER$(\Gamma, \Theta)$}   
           & \multicolumn{9}{c|}{$\text{AER}(\alpha, \varepsilon)$}\\
\hline 
\multirow{5}*{\rotatebox{90}{Mnist}} & $\Gamma$ / $\alpha$           
             & 0.3 & 0.03
             & 1 &1 &1
             &0.3 &0.3 &0.3 
             &0.03&0.03&0.03\\
                                     &$\Theta$ /  $\varepsilon$      
             & 0.05 & 0.1
             &0 &0.05&0.1
             &0 &0.05&0.1
             &0 &0.05&0.1\\
\cline{2-13}
&NP 
    &88.2 & 54.0
    &\textbf{\textcolor{red}{99.4}}   &82.6 &31.1  
    &\textbf{\textcolor{red}{99.3}} &42.2 &0.7
    &\textbf{\textcolor{red}{98.2}} &10.5 &3.0 \\
&AP   
    & \textcolor{blue}{\underline{96.9}}& \textcolor{blue}{\underline{90.7}}
    &98.9  &\textcolor{blue}{\underline{94.0}} &\textcolor{blue}{\underline{77.9}}  
    &98.6 &\textcolor{blue}{\underline{92.3}} &\textcolor{blue}{\underline{74.4}}
    &96.4 &\textcolor{blue}{\underline{85.3}} &\textcolor{blue}{\underline{57.1}} \\
&BAP 
    &\textbf{\textcolor{red}{97.8}} &\textbf{\textcolor{red}{92.6}}
    &\textcolor{blue}{\underline{99.3}}   &\textbf{\textcolor{red}{94.8}} &\textbf{\textcolor{red}{81.2}}  
    &\textbf{\textcolor{red}{99.3}} &\textbf{\textcolor{red}{94.5}} &\textbf{\textcolor{red}{78.7}}
    &\textcolor{blue}{\underline{97.1}} &\textbf{\textcolor{red}{86.5}} & \textbf{\textcolor{red}{58.3}}  \\
\hline
\multirow{4}*{\rotatebox{90}{Cifar10}} & $\Gamma$ / $\alpha$          
             & 0.3 & 0.03
             & 1 &1 &1
             &0.3 &0.3 &0.3 
             &0.03&0.03&0.03\\
                                     &$\Theta$ /  $\varepsilon$  
             &0.01 &0.02
             &0   &0.01 &0.02
             &0 &0.01 &0.02
             &0 &0.01 &0.02\\
\cline{2-13}
&NP 
    & 56.1& 38.0
    &\textbf{\textcolor{red}{90.1}}     &56.7 &37.1 
    &\textbf{\textcolor{red}{87.7}} &24.5 &9.8
    &\textbf{\textcolor{red}{77.8}} &21.6 &13.2 \\
&BAP       
    & \textbf{\textcolor{red}{72.0}}& \textbf{\textcolor{red}{60.7}}
    &82.3 &\textbf{\textcolor{red}{63.9}} &\textbf{\textcolor{red}{44.8}} 
    &82.4 &\textbf{\textcolor{red}{61.0}} &\textbf{\textcolor{red}{38.0}} 
    &77.6 &\textbf{\textcolor{red}{48.0}} &\textbf{\textcolor{red}{23.8}} \\
\hline

\multirow{4}*{\rotatebox{90}{Cifar100}} & $\Gamma$ / $\alpha$        
             & 0.3 & 0.03
             & 1  &1 &1
             &0.3 &0.3 &0.3 
             &0.03&0.03&0.03\\
                                     &$\Theta$ /  $\varepsilon$     
             &0.002 &0.005
             &0 &0.002 &0.005
             &0 &0.002 &0.005
             &0 &0.002 &0.005\\
\cline{2-13}
&NP  
    & \textbf{\textcolor{red}{49.1}}& 35.7
    &\textbf{\textcolor{red}{60.4}}    &35.8 &18.4  
    &\textbf{\textcolor{red}{56.0}} &38.9 &21.1
    &36.4 &31.2 &22.5 \\
&BAP    
    & 48.7& \textbf{\textcolor{red}{40.9}}
    &50.7 &\textbf{\textcolor{red}{43.9}} &\textbf{\textcolor{red}{31.2}} 
    &52.0 &\textbf{\textcolor{red}{44.9}} &\textbf{\textcolor{red}{30.8}}
    &\textbf{\textcolor{red}{44.3}} &\textbf{\textcolor{red}{38.9}} &\textbf{\textcolor{red}{28.0}} \\ 
\hline
\end{tabular}
}
\end{table*} 

We analyze and compare BAP from the point of view of stability. 
First, the stability of BAP results under different attack budgets and compression ratios is compared in Figure~\ref{fig:BAT}.
Similarly to the previous analysis, clean accuracy exhibits the best stability, and the accuracy under attack shows slightly greater instability. However, the overall standard deviation is relatively small and smooth.
Comparing the stability of BAP with that of other methods, we can summarize that the fluctuation of BAP results is not much different from that of the clean accuracy or accuracy of trained budget (AP) (especially for DeepFool attacks with budgets less than $0.1$) and is significantly less than that of the accuracy under untrained attack budgets (AP).
Since the AP is trained directly using clean data and AEs with the prescribed budget, its fluctuation of trained data is small.
Due to BAP's use of the dynamically estimated budget of AEs, AEs with different budgets may be used during training, and the stability of the calculation results can be guaranteed for all budgets. BAP thus exhibits better stability than existing AP methods. 
Therefore, BAP can achieve better overall AER performance (taking accuracy, efficiency and robustness into account) with more stable results.

\section{Conclusion and Future Work}

Most existing studies only consider efficiency or robustness alone; fewer research efforts combine accuracy, efficiency and robustness (AER).
In this paper, we analyze the performance of the model under different compression ratios and attack budgets for the gradual pruning process, finding that different pruning ratios drastically impact robustness. 
The robustness of the existing NP approaches drastically varies with different pruning processes, especially under attacks with large strength.
Few AP approaches exist that combine efficiency and robustness by mixing the clean data and AEs with prescribed budgets into the gradual pruning process; these approaches cannot obtain the models with the best comprehensive performance, and the robustness of existing models exhibits high sensitivity to the budget and is only robust when confronting attacks with the same budget.
Furthermore, to better balance the AER between the different training approaches, this paper proposes a training method based on model pruning and blind adversarial training called blind adversarial pruning (BAP). 
The main idea of this approach is to use BAT to adaptively estimate a nonuniform budget to modify the AEs used in the training, ensuring that the strengths of the AEs are dynamically located within a reasonable range, and to work with pruning to ultimately improve the overall AER of the pruned model. 
The experimental results obtained using BAP for pruning classification models on several benchmarks demonstrate the competitive performance of this method: the robustness of the pruned model by BAP is more stable with varying pruning processes; and BAP exhibits better overall AER than AP does. 

The present research is still in the early stage. There are several aspects that deserve deeper investigation: 
\begin{itemize}
  \item The effectiveness of the BAP algorithm could be verified through richer numerical experiments, especially for more complex datasets.
  \item Theoretical analysis could be combined to discuss the theoretical commonalities of the AER of different models.
  \item The effects and improvement of different training parameters with respect to the results of BAP could be studied in greater detail.
  \item The idea of BAP could be applied to more abundant compression and adversarial methods to find the essential advantages and disadvantages of the method.
\end{itemize}

\section*{Acknowledgements}

This work was supported in part by the Innovation Foundation of Qian Xuesen Laboratory of Space Technology, and in part by Beijing Nova Program of Science and Technology under Grant Z191100001119129. 

\bibliographystyle{ieee_fullname}
\bibliography{paper_ref,paper_ref_comp}
\end{document}